\documentclass[11pt]{article}

\usepackage[preprint]{acl}

\usepackage{times}
\usepackage{latexsym}
\usepackage[T1]{fontenc}

\usepackage[utf8]{inputenc}

\usepackage{microtype}

\usepackage{inconsolata}

\usepackage{graphicx}
\usepackage{amsmath}
\usepackage{booktabs}

\title{When Bigger Becomes Brittle:\\ Why Compression Hurts More as Language Models Scale}

\author{
  \small
  \textbf{Gautam Gupta\textsuperscript{1,$*$}},
  \textbf{Dikshant Kukreja\textsuperscript{2,$*$}},
  \textbf{Unique Kumar\textsuperscript{1}},
  \textbf{Avni Mittal\textsuperscript{3}},
  \textbf{Deepanshu Dabas\textsuperscript{4}},
\\
  \small
  \textbf{Kshitij Sah\textsuperscript{1}},
  \textbf{Erik Cambria\textsuperscript{5}},
  \textbf{Rajiv Ratn Shah\textsuperscript{6}},
  \textbf{Zhengkui Wang\textsuperscript{7}},
  \textbf{Avinash Anand\textsuperscript{7}}
\\[1pt]
  \small
  \textsuperscript{1}Indraprastha Institute of Information Technology, Delhi \quad
  \textsuperscript{2}TurboML
\\[-2pt]
  \small
  \textsuperscript{3}Independent Researcher \quad
  \textsuperscript{4}Microsoft \quad
  \textsuperscript{5}Nanyang Technological University
\\[-2pt]
  \small
  \textsuperscript{6}Indian Institute of Technology, Kanpur \quad
  \textsuperscript{7}Singapore Institute of Technology
}

\begin{document}
\maketitle
\begingroup\renewcommand{\thefootnote}{$*$}\footnotetext{Equal contribution}\endgroup
\begin{abstract}

Quantization and pruning reduce the cost of deploying language
models, but their effects are usually reported as benchmark scores.
Such scores record only the loss that remains at the output. Inside
the model, compression damages some components more than others, and
the rest of the network can partly compensate. The observed loss
therefore reflects both the damage and this compensation, which
benchmark scores cannot separate. How the two combine, and whether
their balance changes with model size, remains unclear. We study this
compensation for in-context retrieval, a capability carried by a
small, identifiable set of attention heads. To separate the two, the
expected loss from ablating these heads is decomposed exactly into the
part restored by the rest of the network, attributed to the
components responsible, and the part that reaches the output. This
decomposition is then compared across model sizes. For damage of
comparable size, smaller models restore most of the loss, whereas
larger models retain most of it, because the compensating responses
offset less of it as models grow. Compressing the heads produces the
same pattern. Compression that is nearly harmless in small models can
therefore be costly in large ones.

\end{abstract}

\section{Introduction}

Quantization and pruning reduce language-model deployment costs by lowering weight precision or removing parameters \citep{frantar2023gptq, lin2024awq, frantar2023sparsegpt, sun2024wanda}. Their effects, however, are not uniform. Earlier work on neural-network compression reports disproportionate degradation on rare examples \citep{hooker2019forget}, while studies of language models identify losses in knowledge-intensive tasks \citep{namburi2023cost, jaiswal2024llmkick}, long-context use \citep{mekala2025quantization}, and lower-resource languages \citep{marchisio2024multilingual}. Aggregate metrics can obscure these changes. \citet{jaiswal2024llmkick} show that perplexity can miss failures on knowledge-intensive tasks, while \citet{dutta2024accuracy} show that similar accuracy can conceal changes from correct to incorrect answers and vice versa. Preserving an overall score therefore does not establish that individual capabilities or predictions have been preserved.

Internal measurements provide a complementary view of what compression preserves. \citet{singh2025interpreting} report generally minor calibration changes and stable counts of inactive neurons under 4-bit and 8-bit quantization. Such stability, however, does not establish that individual features remain intact. Using a fixed sparse autoencoder to compare activations, \citet{duan2026sae} finds that INT7 quantization of Gemma-2-2B degrades 18.7\% of active features despite improving perplexity. A related distinction appears in image classification: \citet{rabbi2026explanations} find that competitive classification accuracy in INT8-quantized EfficientNet-B0 can coexist with degraded input-feature attributions. These results illustrate why output scores and coarse internal summaries provide an incomplete account of compression-induced changes.

Component-level analyses further examine where these changes arise. Outlier features can make low-precision computation difficult \citep{dettmers2022llmint8, sun2024massive}, while attribution analyses trace quantization-related losses in factual recall to changes in particular neurons and layers \citep{wang2025compressed}. At lower precision, \citet{zhou2026failure} distinguish signal degradation, in which computational patterns remain intact but information loses precision, from computation collapse, in which essential components cease to function. Identifying affected computations can also guide compression: Task-Circuit Quantization preserves task-relevant weights at higher precision while quantizing the remaining weights \citep{xiao2025tacq}. Together, these studies connect compression-induced changes to the computations that support a task.

Damage to a task-relevant component need not produce a corresponding reduction at the output: other components can partly compensate, as shown by the Hydra effect and self-repair \citep{mcgrath2023hydra, rushing2024selfrepair}. We ask how much this response offsets comparable circuit damage across model sizes, measuring both the contribution removed and the response of the remaining network.

We study in-context retrieval using indirect-object identification (IOI; \citealp{wang2022ioi}). In ``When Mary and John went to the store, John gave a drink to \ldots'', the model must retrieve ``Mary'' rather than repeat ``John.'' Retrieval heads attend to the answer name and increase its logit, while copy-suppression heads reduce that logit \citep{mcdougall2023copy}. After retrieval-head damage, we account exactly for the gap between the removed direct contribution and the measured output reduction through changes in suppression, other components' contributions, and final normalization. Across the scaling suite, larger models restore a smaller share of comparable retrieval deficits, although selected backup responses remain protective at every model size.

To examine how this distinction applies to compression, we repeat the accounting under retrieval-head quantization and structural head pruning. Quantization alters the targeted heads' weights, whereas pruning can remove both retrieval contributions and components that would otherwise participate in repair. The replications also show reductions in net repair with model size, although their intervention scopes differ. These results identify a mechanism through which compression tolerance can depend on scale: the output effect of damaged computation also depends on the network response that follows it. Together, our analyses connect circuit-level damage to output behavior and identify a scale-dependent limit on net repair in the studied retrieval circuit. Apparent tolerance to compression at one scale is therefore not sufficient evidence of comparable tolerance at another.

\section{Methodology}
\label{sec:method}

\paragraph{Models.}
For single-model analyses, we use GPT-2 XL \citep{radford2019language} and Llama 3.1 8B \citep{grattafiori2024llama} to examine repair mechanisms in two distinct model families. For scaling analyses, we use Pythia \citep{biderman2023pythia} as our primary suite. Chinchilla scaling laws describe how language-model loss depends on both parameter count and training-token count \citep{hoffmann2022training}, motivating comparisons that control training exposure. Pythia supports this comparison because its models are trained on the same data, in the same order, with a fixed training-token count. We evaluate the final checkpoints of six Pythia models: 410M, 1B, 1.4B, 2.8B, 6.9B, and 12B parameters. We additionally verify the observed scaling trend in the OPT family \citep{zhang2022opt}, testing whether it extends beyond Pythia. All models are loaded using TransformerLens \citep{nanda2022transformerlens}, and all measurements are performed in fp32.

\paragraph{Data.}
We use 1{,}000 exploratory IOI prompts and a separate set of 2{,}000 held-out prompts, all using single-token names. Each set is shared across models. Target selection, damage calibration, and backup-receiver selection are fixed on the exploratory prompts before held-out evaluation.

\paragraph{Setup.}
Retrieval heads are selected using the IOI-adapted score of \citet{wu2024retrieval} at a threshold of 0.1. Copy-suppression heads are identified using the OV and QK circuit tests of \citet{mcdougall2023copy}, with analysis-specific selection rules in Appendix~\ref{sec:app-setup}. Ablation sweeps zero the outputs of the top-$k$ retrieval heads at all positions, ranked by clean direct logit attribution and excluding copy-suppression heads \citep{elhage2021mathematical}. The backup response is the net change in direct contributions from all heads and MLP layers outside the target and copy-suppression sets \citep{rushing2024selfrepair}.

\paragraph{Quantization and pruning.}
We use simulated per-channel weight quantization \citep{nagel2021white} and attention-head masking \citep{michel2019sixteen} as compression interventions. We quantize the target retrieval heads' attention-projection weights to 4-bit integers per output channel and dequantize them in place for fp32 inference, leaving other parameters unchanged. For pruning, structural masks zero half the eligible attention heads at all positions, using typical and ``unlucky'' masks (three seeds each). Copy-suppression heads are excluded from both interventions, and we measure damage and repair using the same accounting as ablation (Appendix~\ref{sec:app-setup}).

\section{Tracing the Damage Response}
\label{sec:mechanism}

For each IOI prompt, let $z_{\mathrm{ans}}$ and $z_{\mathrm{subj}}$
denote the logits of the correct name and the repeated subject name
at the final prediction position. We measure the logit difference
\[
  m = z_{\mathrm{ans}} - z_{\mathrm{subj}}.
\]
Let $a_i$ denote component $i$'s direct logit attribution to this
difference, evaluated using the same clean-run final-normalization
scale in both clean and intervened runs
(\citealp{elhage2021mathematical}; Appendix~\ref{sec:app-setup}).
We write $\Delta q = q^{\mathrm{int}} - q^{\mathrm{clean}}$ for
intervention-induced changes.

We partition the set $\mathcal{C}$ of attention heads and MLP layers
into the targeted heads $T$, the copy-suppression heads
$\mathrm{CS}$, and the remaining components
$O = \mathcal{C} \setminus (T \cup \mathrm{CS})$, and define
\[
\begin{alignedat}{2}
  D &= -\textstyle\sum_{i\in T}\Delta a_i
    &\quad& \text{(deficit)},\\[3pt]
  H &= -\Delta m
    && \text{(output harm)},\\[3pt]
  R &= \textstyle\sum_{i\in\mathrm{CS}}\Delta a_i
    && \text{(CS release)},\\[3pt]
  B &= \textstyle\sum_{i\in O}\Delta a_i
    && \text{(net backup response)},\\[3pt]
  N &= \Delta\bigl(m-\textstyle\sum_{i\in\mathcal{C}}a_i\bigr)
    && \text{(final normalization)}.
\end{alignedat}
\]
Here, $N$ captures the change in the part of the score that the
component contributions do not account for under the fixed
clean-scale readout. Under zero-ablation, $a_i^{\mathrm{int}}=0$ for
$i\in T$, so the deficit reduces to
$D=\sum_{i\in T}a_i^{\mathrm{clean}}$. These definitions yield the
accounting identity
\begin{equation}
  H = D - R - B - N.
  \label{eq:accounting}
\end{equation}
Total repair, $D-H=R+B+N$, is the signed difference between the
measured score $m^{\mathrm{int}}$ and the no-response baseline
$m^{\mathrm{clean}}-D$. Positive values of $R$, $B$, and $N$ offset
the deficit, whereas negative values increase harm. Positive $H$
indicates a reduced advantage for the correct name, without
necessarily changing which of the two names has the higher logit.

\begin{figure}[t]
  \centering
  \includegraphics[width=\columnwidth]{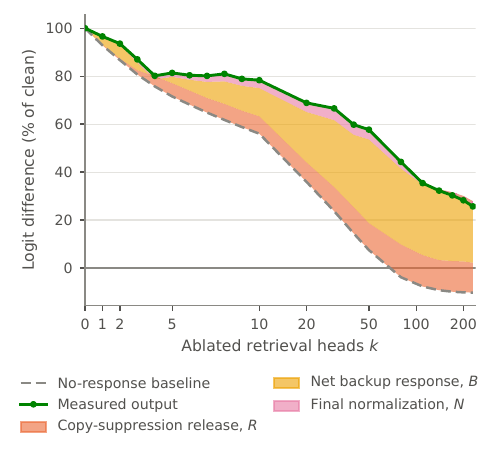}
  \caption{Repair decomposition in GPT-2 XL as the top-$k$
  retrieval heads are ablated. The gap between the no-response
  baseline and the measured output is filled by copy-suppression
  release, the net backup response, and final normalization.
  Values are percentages of the clean logit difference.}
  \label{fig:gpt2xl-bands}
\end{figure}

On an undamaged run, retrieval heads attend from the final prediction
position to the answer name in context and write a contribution to
the residual stream that increases its output logit
\citep{wang2022ioi}. Later copy-suppression heads act on this
retrieved signal, attending back to the predicted name and reducing
its logit \citep{mcdougall2023copy}. Their clean-run contribution
defines the \emph{standing suppression},
\[
  S=-\sum_{i\in\mathrm{CS}}a_i^{\mathrm{clean}},
  \qquad
  S^{\mathrm{int}}=S-R.
\]
Thus, positive $R$ measures a reduction in suppression relative to
the clean run. The backup response $B$ similarly measures the net
change from the clean-run contributions of all remaining heads and
MLP layers, including surviving retrieval heads, rather than only
positively responding components. All quantities are computed per
prompt and reported as means over prompts.

\subsection{Single-Model Analysis}
\label{sec:single}

We examine the response to retrieval-head damage in GPT-2 XL by
progressively zeroing the outputs of the top-$k$ retrieval heads at
all token positions. Heads are ranked by their clean direct
contribution to the answer--distractor logit difference, excluding
copy-suppression heads (Section~\ref{sec:method}).

Under zero-ablation, $D$ is the summed clean contribution of the
removed heads. Holding all remaining contributions and the
final-normalization scale fixed would give the no-response baseline,
$m^{\mathrm{clean}}-D$. The measured output instead remains above
this baseline (Figure~\ref{fig:gpt2xl-bands}). At the largest
displayed settings, the mean answer margin remains positive even
though the baseline falls below zero. We explain this partial
compensation through two component-response channels and a separate
final-normalization term.

\paragraph{Copy-suppression release.}
The first channel offsets damage by reducing a negative contribution
already present on the clean run. Copy-suppression heads act on the
retrieved signal and reduce support for the answer. Their targeting
depends on that signal: when the retrieval heads' contribution to the
suppressors' input is replaced with a baseline, the suppressors
attend about equally to the answer and the distractor
\citep{mcdougall2023copy}. When their suppression weakens after
retrieval-head damage, the answer margin increases relative to the
no-response baseline. The released contribution is
$R=S-S^{\mathrm{int}}$, where $S$ and $S^{\mathrm{int}}$ are the
clean and intervened suppression. Positive $R$ therefore means that
the suppressors subtract less, not that the ablated heads recover
their output. If suppression approaches zero, release approaches the
standing suppression $S$. In GPT-2 XL, the identified heads apply
0.41 logits of standing suppression, about 9\% of the clean logit
difference, and ablating the ten highest-ranked retrieval heads
releases 0.32 logits of it.

\paragraph{The backup response.}
The second channel is the signed net change in contributions from
all heads and MLP layers outside the targeted and copy-suppression
sets, including surviving retrieval heads. Related increases in
downstream attention contributions were described as the Hydra
effect \citep{mcgrath2023hydra}. Positive $B$ means that these
remaining components collectively contribute more to the answer
margin after damage. Because individual changes can oppose one
another, $B$ measures net restoration rather than the magnitude of
all adjustments. In GPT-2 XL, this response accounts for most of the
repair gap at larger ablation settings: with ten heads ablated, it
supplies 0.51 of the 0.97 logits of repair, and from twenty heads
onward it is at least twice the release term
(Figure~\ref{fig:gpt2xl-bands}).

\begin{figure}[t]
  \centering
  \includegraphics[width=\columnwidth]{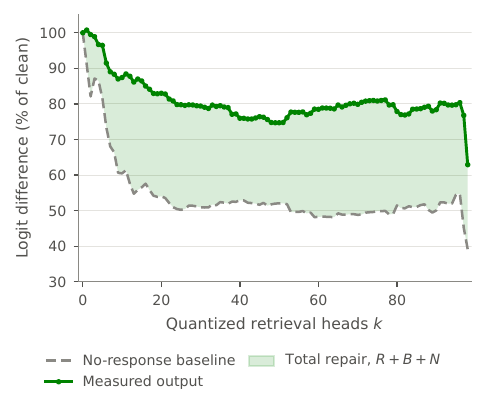}
  \caption{Repair under 4-bit weight quantization of the top-$k$
  retrieval heads in Pythia-410M, using the deliberately poor
  calibration described in Appendix~\ref{sec:app-setup}. The shaded
  gap between the no-response baseline and the measured output is
  total repair, $R+B+N$. Values are percentages of the clean logit
  difference.}
  \label{fig:quantgap}
\end{figure}

\paragraph{Final normalization.}
Component attributions use the same clean-run final-normalization
scale, whereas the actual intervened output uses the normalization of
the intervened residual stream. The term $N$ accounts for the
resulting difference not captured by the fixed-scale contributions.
It is a normalization term, not a third population of compensating
components, and is comparatively small in GPT-2 XL: it reaches at
most 0.21 logits and turns slightly negative beyond about 140
ablated heads, where it adds to the harm. Together, the two response
channels and normalization account for total repair,
$D-H=R+B+N$ (Equation~\ref{eq:accounting}), with a residual below
$4\times10^{-3}$ logits in GPT-2 XL that arises from the
final-normalization bias folded into the unembedding. Their
contributions can be positive or negative, respectively offsetting
the deficit or increasing harm.

\paragraph{From ablation to quantization.}
In a separate experiment on Pythia-410M, we quantize progressively
larger sets of top-ranked retrieval heads. Their four attention
projections are quantized to 4-bit integers per output channel and
dequantized for fp32 inference, leaving biases and other parameters
unchanged and excluding copy-suppression heads. Unlike ablation,
quantization leaves the targeted heads active; $D$ is therefore their
measured reduction in direct contribution under the clean-run
normalization scale.

The measured output again lies substantially above the no-response
baseline at every quantization setting (Figure~\ref{fig:quantgap}).
With ten quantized heads, for example, the targeted heads lose 1.70
logits of direct contribution, yet the output falls by only 0.54
logits. The shaded gap is total repair, $R+B+N$, showing that
compensation is not confined to zero-ablation. This sweep uses
deliberately poor calibration (Appendix~\ref{sec:app-setup}), so it
illustrates compensation under a specified weight perturbation rather
than the performance of an optimized deployment quantizer.

These experiments distinguish the direct-contribution deficit from
the harm remaining at the output. We next examine how much comparable
retrieval-head damage is offset across model sizes and how the two
response channels contribute to this balance.

\begin{figure*}[t]
  \centering
  \includegraphics[width=\textwidth]{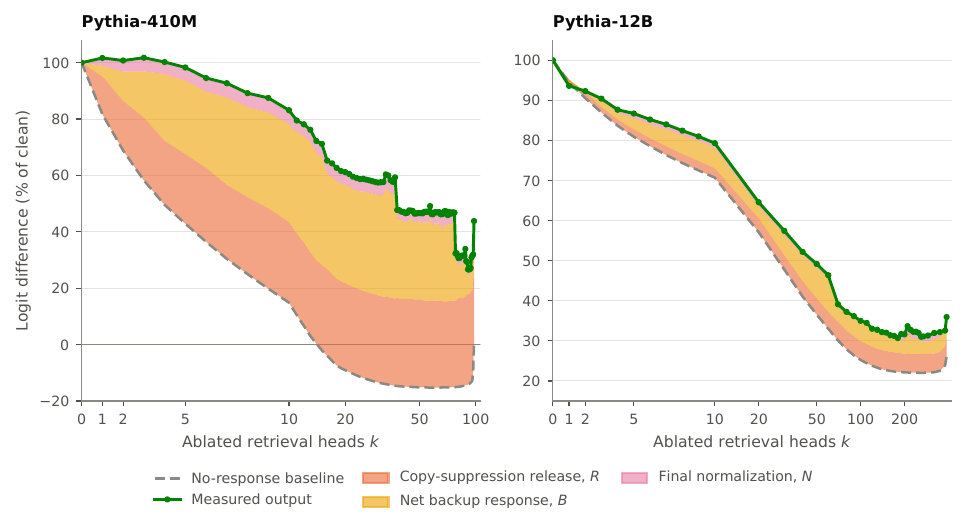}
  \caption{Repair decomposition at the two ends of the Pythia suite,
  Pythia-410M (left) and Pythia-12B (right), as the top-$k$ retrieval
  heads are ablated. The gap between the no-response baseline and the
  measured output is filled by copy-suppression release, the net backup
  response, and final normalization, which together account for it to
  within $10^{-3}$ logits. At 410M the response offsets most of the
  removed contribution; at 12B it offsets a much smaller share. Values
  are percentages of each model's clean logit difference.}
  \label{fig:c2bands}
\end{figure*}

\subsection{Scaling Analysis}
\label{sec:scale}

\begin{figure*}[t]
  \begin{minipage}[t]{0.48\textwidth}
    \centering
    \includegraphics[width=\linewidth]{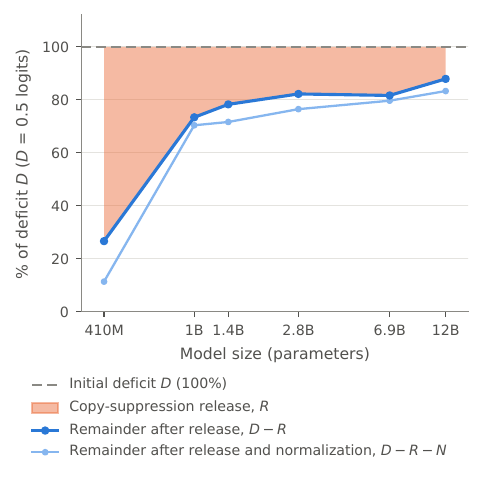}
    \caption{Deficit remaining after copy-suppression release at
  $D=0.5$ logits, estimated by interpolation within each model's
  ablation sweep. The shaded region between the initial deficit
  (dashed line) and the remainder $D-R$ (solid line) is
  copy-suppression release; output harm $H$ also depends on the
  backup response and final normalization. Values are percentages
  of $D$.}
    \label{fig:c3harm}
  \end{minipage}\hfill
  \begin{minipage}[t]{0.48\textwidth}
    \centering
    \includegraphics[width=\linewidth]{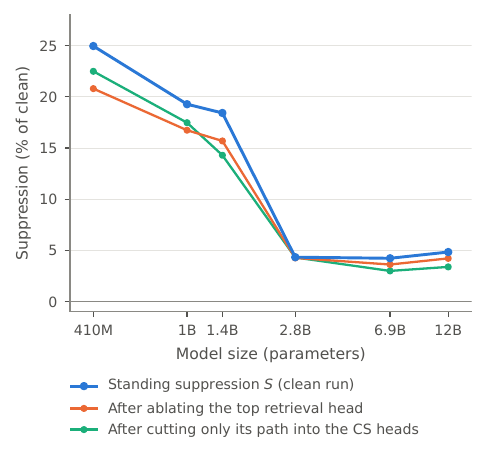}
    \caption{Standing suppression $S$ across scale, measured on the clean
  run, as a percentage of each model's clean logit difference.
  Less standing suppression in the larger models leaves a
  smaller contribution available for release after damage.}
    \label{fig:c3cs}
  \end{minipage}
\end{figure*}

We examine how the response to retrieval-head damage varies with
model size by progressively ablating the top-$k$ retrieval heads in
each of the six Pythia models. The ten highest-ranked heads
contribute different amounts across models, so ablating the same
number of heads does not produce the same deficit. We therefore
compare models at a shared deficit of $D=0.5$ logits, estimated by
linear interpolation between neighboring settings of each model's
ablation sweep.

At the shared deficit of $D=0.5$ logits, holding all remaining contributions and the final-normalization scale fixed would reduce the answer--distractor logit difference by 0.5 logits. This gives the no-response baseline. We estimate the actual output at the same deficit by interpolating each model's final logit difference between neighboring ablation settings, then comparing it with the clean-run value. At 410M, copy-suppression release, the backup response, and final normalization together provide about 0.55 logits of repair. This offsets the 0.5-logit deficit and leaves the output margin about 0.05 logits above its clean value. At 12B, about 0.38 of the original 0.5 logits remains as output harm. We next examine how each response channel contributes to this difference.

\paragraph{Copy-suppression release.}
Release offsets most of the deficit at 410M but little of it at
12B: the remainder after release, $D-R$, is about 27\% of $D$ at
410M, compared with nearly 90\% at 12B, and this fraction increases
overall with model size (Figure~\ref{fig:c3harm}). The
copy-suppression heads apply about 1.1 logits of standing
suppression at 410M, about 25\% of the clean logit difference,
compared with 0.2 logits, about 5\%, at 12B
(Figure~\ref{fig:c3cs}). The greater standing suppression at 410M
leaves more that can be released to compensate for retrieval-head
damage.

\begin{figure}[t]
  \centering
  \includegraphics[width=\columnwidth]{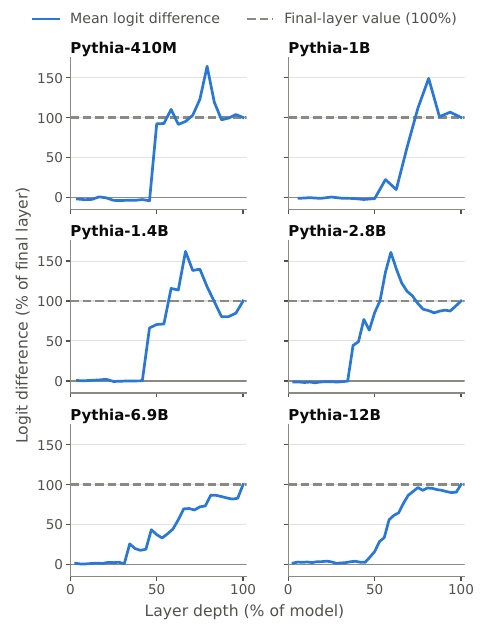}
  \caption{Mean clean-run answer--distractor logit difference across
  layer depth, with the intermediate residual normalized separately
  at each layer, as a percentage of each model's final-layer value
  (dashed line at 100\%). Several models show an intermediate peak
  above the final value; the 6.9B and 12B curves reach their maxima
  at the final layer.}
  \label{fig:c3lens}
\end{figure}

To examine how the models form the answer, we read the clean
answer--distractor logit difference after each layer with the
logit lens \citep{nostalgebraist2020logitlens,belrose2023tuned}.
In the four smaller models, the mean difference first rises above
its final value and then falls. We call this rise an
\emph{overshoot}: the intermediate prediction favors the answer
over the distractor more strongly than the final prediction does.
At 6.9B and 12B, the mean difference instead reaches its maximum at
the output (Figure~\ref{fig:c3lens}). We hypothesize that these
differences reflect increasing specialization in answer formation.
In this account, smaller models rely more on later suppression to
reduce intermediate overshoot, whereas larger models form the
answer with less need for this correction. This would explain both
the greater standing suppression in smaller models and the larger
contribution its release can make to repair after retrieval-head
damage.

To relate changes in intermediate predictions to changes in suppression within each model, we evaluated Pythia-410M, 1B, 1.4B, and 6.9B at several stages of training on the held-out prompts. At each checkpoint, we read the clean logit difference after each layer with the same logit lens and average it across prompts. The \emph{intermediate excess}, the
size of the overshoot, is the amount by which the maximum of this
mean trajectory exceeds that checkpoint's own final-layer value.
We also identify copy-suppression heads separately at each
checkpoint and measure their standing suppression on the same clean
runs (Appendix~\ref{sec:app-training-codevelopment}). Standing
suppression broadly follows the developmental profile of the
intermediate excess. The training histories include periods in
which increasing excess is accompanied by stronger standing
suppression. They also include stabilization or decline after an
earlier rise, and cases in which both quantities remain weak
(Appendix~Figure~\ref{fig:app-training-codevelopment}). Because the
checkpoints are observed rather than manipulated, this
co-development is consistent with the hypothesis that suppression
develops to correct the excess but does not establish it.

\begin{figure}[t]
  \centering
  \includegraphics[width=\columnwidth]{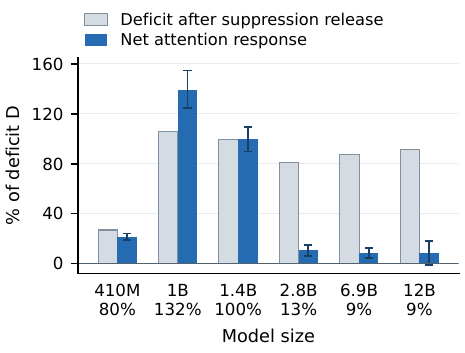}
  \caption{Net attention response after comparable retrieval-head
damage. Gray bars show the deficit left after copy-suppression
release, $D-R$, and blue bars show the net attention response.
Bar heights are percentages of $D$; labels below the models give
the percentage of $D-R$ restored. Error bars show 95\% bootstrap
intervals.}
  \label{fig:c3response}
\end{figure}

\paragraph{The backup response.}
The attention component of the backup response offsets much of
the deficit left after copy-suppression release in smaller models,
but only a small fraction in larger models. To compare models at
similar retrieval-head deficits, we calibrate the ablations on
exploratory prompts to a deficit of about 0.8 logits, then evaluate
the fixed interventions on held-out prompts. We measure the
attention response by summing signed changes in direct
contributions from heads outside the targeted and
copy-suppression sets.

At 410M, the net attention response offsets about 80\% of the
remainder $D-R$, compared with about 9\% at 12B
(Figure~\ref{fig:c3response}). In the three smaller models, this
response is comparable to the remainder after release. From 2.8B
upward, it offsets only about one tenth of that remainder. The
remainder itself is similar from 1B to 12B, so the smaller net
response does not reflect a smaller amount left to repair.

A small net attention response leaves open how much the responding
heads affect the final prediction.
To measure this effect, we compare the final answer--distractor
logit difference with selected adjustments allowed and with them prevented.
We hold selected responding heads at their clean-run output at the
answer position, allowing subsequent computation to run.
Applying the same intervention to heads with little response
provides a control.

Preventing the selected responders' adjustment increases harm in every model.
At 12B, it increases harm by about 0.30 logits despite a net
attention response of only 0.07 logits
(Table~\ref{tab:backup-harm};
Appendix~Figure~\ref{fig:c3protect}).
The harm increase includes downstream effects of preventing
selected adjustments; the net attention response sums signed
changes across all eligible remaining heads.
Preventing both backup and copy-suppression adjustments raises
harm beyond the original deficit in the three largest models
(Table~\ref{tab:backup-harm}).
Thus, larger models retain protective backup responses, even
though the net attention response provides little restoration of
the answer margin.

The set selected to cover 90\% of the positive response on exploratory
prompts comprises one head at 1B and 34 at 12B.
One possibility is that a response distributed across more heads
includes opposing changes that partly cancel in the net sum.
Testing whether such cancellation explains the scale difference
remains future work.

\begin{table}[!b]
  \centering
  \small
  \setlength{\tabcolsep}{4pt}
  \renewcommand{\arraystretch}{1.1}
  \begin{tabular}{@{}lrrrr@{}}
    \toprule
          & & \multicolumn{3}{c@{}}{Output harm $H$} \\
    \cmidrule(l){3-5}
    Model & Deficit $D$ & Natural &
    \shortstack{Backup\\held clean} & \shortstack{Both\\held clean} \\
    \midrule
    410M & 0.797 & $-0.067$ & 0.078    & 0.728 \\
    1B   & 0.795 & $-0.226$ & $-0.125$ & 0.815 \\
    1.4B & 0.756 & $-0.027$ & 0.354    & 0.464 \\
    2.8B & 0.888 & 0.476    & 1.316    & 1.770 \\
    6.9B & 0.829 & 0.561    & 0.675    & 0.905 \\
    12B  & 0.851 & 0.581    & 0.878    & 1.018 \\
    \bottomrule
  \end{tabular}
  \caption{Output harm after retrieval-head damage with natural
responses, selected backup heads held at their clean outputs, or
both backup and reference copy-suppression heads held clean.
Values are means in logits; negative $H$ indicates an answer
margin above its clean value. $D$ gives the natural-condition
deficit; at 1.4B, both prevention conditions reduce it to
0.707 logits.}
  \label{tab:backup-harm}
\end{table}

\paragraph{From ablation to compression.}
The scale contrast also appears under 4-bit retrieval-head
quantization. Output harm as a fraction of the deficit, $H/D$,
rises overall from 0.55 at 410M to 0.91 at 12B in Pythia, with
variation at intermediate sizes (Appendix~\ref{sec:appendix}).
Under structural pruning, mean repair decreases with size for
unlucky masks while remaining positive; random-mask repair is
non-monotonic (Appendix~\ref{sec:appendix-prune}).

Together, these analyses connect the scale difference in output
harm to changes in both response channels.
Smaller models can offset much of the removed contribution through
copy-suppression release and the backup response; larger models retain
a greater share as output harm.
For suppression, reduced repair accompanies less standing
suppression and is consistent with the overshoot hypothesis.
For the backup response, protective adjustments remain present,
but the net attention response restores less of the lost margin.

\section{Conclusion}
\label{sec:conclusion}

The single-model analysis shows why a small output change under compression need not imply that the affected computation is intact. Copy-suppression release and changes in the remaining components can partly offset the lost retrieval contribution; final normalization accounts for an additional difference at the output. The quantization experiment demonstrates the same distinction while the targeted heads remain active: their direct contribution can fall substantially even when the output changes much less. In this quantization setting, apparent tolerance therefore depends partly on compensation elsewhere in the network.

The scaling analysis shows that this compensation cannot be assumed to remain equally effective as models grow. Larger models offset less of comparable retrieval-head damage, with less suppression available for release and a smaller net attention response, although protective backup responses remain active. Under quantization, the fraction of the deficit reaching the output also increases overall with model size in both families. Under pruning, mean repair remains positive but declines with size for unlucky masks; random-mask repair is non-monotonic. This suggests that an unlucky compression pattern disproportionately affecting heads supporting one capability could leave it more impaired than others, particularly when compensation is limited. Tolerance in a small model therefore does not establish comparable tolerance at a larger scale. Whether this explains uneven losses across capabilities under practical compression remains to be tested.

\section*{Limitations}

The analysis focuses on in-context retrieval on templated IOI
prompts. Extending this approach to other capabilities and model
families would test the generality of the observed repair patterns.

The checkpoint study tracks the co-development of intermediate
excess and copy suppression. Interventions during training could
establish whether this association reflects a causal relationship.
Comparisons across normalization conventions and answer scores
show that the measured excess depends on the readout
(Appendix~Figures~\ref{fig:app-lens-normalization}
and~\ref{fig:app-lens-vocabulary}).
Further analysis of opposing head contributions could test
whether cancellation explains the decline in net backup
restoration with scale.

Quantization and pruning are studied through dequantized
weights and structural masks in floating-point inference.
These interventions preserve copy-suppression heads, so repair
may differ when compression also damages these components.
Evaluation with deployment kernels would extend these findings
to practical implementations.

\bibliography{custom}

\appendix

\section*{Appendix}

\section{Experimental Setup}
\label{sec:app-setup}
\label{sec:appendix-details}

This section gives the experimental setup;
Section~\ref{sec:method} summarizes it.

\paragraph{Models and numerical settings.}
We use the final checkpoints (step 143{,}000) of Pythia-410M, 1B,
1.4B, 2.8B, 6.9B, and 12B \citep{biderman2023pythia}, loaded in
TransformerLens \citep{nanda2022transformerlens}. Layer normalization
is folded into adjacent linear maps, the writing weights and the
unembedding are centered, and value biases are folded into the output
bias. All runs use fp32, and the held-out evaluations also disable
TF32. The held-out evaluations were run with PyTorch 2.11.0 and
TransformerLens 2.18.0.

\paragraph{Prompts.}
All experiments use the IOI task \citep{wang2022ioi} with names that
tokenize to a single token. The exploratory set contains 1{,}000
prompts generated with seed 0 and shared across models. It is used
for the ablation sweeps, the comparison at $D=0.5$, the ten-head
comparisons, and the quantization and pruning replications. The
held-out set contains 2{,}000 fresh, deduplicated prompts generated
with seed 20260908 from the same 30 templates. It is used for the
comparable-damage and response-prevention experiments, the checkpoint
analysis, and the readout comparisons. In these held-out analyses the
exploratory prompts serve as discovery data: target selection, damage
calibration, and receiver selection are fixed on them before the
held-out prompts are evaluated.

\paragraph{Score and direct contributions.}
The score is the logit of the correct name minus the logit of the
distractor at the final position. The direct contribution of a
component is its write to the residual stream projected onto the
difference between the two names' unembedding vectors and divided by
the final layer-normalization scale of the clean run on the same
prompt \citep[direct logit attribution;][]{elhage2021mathematical}.
The same clean scale is used on clean and intervened runs. Pairwise
accuracy is the fraction of prompts on which the correct name has the
higher logit.

\paragraph{Retrieval heads.}
We adapt the retrieval score of \citet{wu2024retrieval} to IOI
prompts. On each prompt, a head scores one half if its strongest
attention from the final position falls on the correct name while the
model's top prediction is that name, and one half if its strongest
attention from the second occurrence of the subject falls on the
first occurrence. Attention to the beginning-of-sequence token is
excluded. Heads whose score, averaged over prompts, is at least 0.1
are retrieval heads. They are ranked by their clean direct
contribution, and copy-suppression heads are removed from the list.

\paragraph{Copy-suppression heads.}
Copy-suppression heads are selected with the static OV and QK tests of
\citet{mcdougall2023copy}, computed on a random sample of 2{,}048
vocabulary tokens. The OV score of a head is the fraction of sampled
tokens whose own entry in the head's OV circuit, mapped from the
effective embedding to the unembedding, is among the five most
negative entries in its column. The QK score is the fraction of
sampled tokens whose own key ranks among the top five when the
token's unembedding is the query. A head's score is the smaller of
the two. The Pythia panels in Figure~\ref{fig:app-repair-bands} use the heads that
score at least 0.5: one head at 410M and two at 12B. The
scaling analysis and the held-out experiments add each model's two
highest-scoring heads to this set, which gives two heads at every
size (Table~\ref{tab:models}); this is the reference set. The
checkpoint analysis instead selects heads functionally, by a negative
clean direct contribution, attention to the correct name, and the OV
and QK tests, with no minimum number of heads. The response-prevention
experiments also report this functional set for comparison.

\paragraph{Backup response and receivers.}
Backup heads are not selected in advance \citep{rushing2024selfrepair}.
The backup response $B$ is the change in the summed direct
contribution of all heads and MLP layers outside the target and
copy-suppression sets. For response prevention, the backup receivers
are the smallest set of heads, outside the targets and the
copy-suppression heads, that covers 90\% of the positive attention
response on the discovery prompts.

\paragraph{Ablation.}
Zero-ablation sets a target head's output to zero at all token
positions. Targets are the top-$k$ retrieval heads. Ablation sweeps
use every $k$ when the retrieval set has at most 100 heads, and
otherwise $k=1, 10, 20, 30, 40, 50$, then steps of 30 up to the full
set. The ten-head comparisons use $k=10$. The comparison at $D=0.5$
interpolates linearly between neighbouring settings of each model's
sweep, using the sweep up to the setting with the largest $D$.

\paragraph{Comparable damage.}
The held-out ablations are calibrated on the discovery prompts to
three deficits of about 0.43, 0.86, and 1.71 logits. Ranked retrieval
heads are zeroed whole when this reaches a target within 10\%.
Otherwise, the live output of the last head in the ranking is scaled
down until the target is reached. The resulting interventions are
frozen and applied unchanged to the held-out prompts, where $D$ is
measured again. The middle level is the primary condition.

\paragraph{Path cut and response prevention.}
The path cut removes the retrieval heads' output only from the input
of the selected downstream heads, so every other component still
receives it. Response prevention holds selected heads at their clean
output: their outputs before the output projection are replaced with
clean-run values at the answer position, while all later computation
runs normally. Separate conditions hold the copy-suppression heads,
the backup receivers, or both. Controls apply the same replacement to
heads with little response and replace clean values with clean
values. $D$ is measured in every condition.

\paragraph{Quantization.}
Quantization is simulated. The weights of each target head's four
attention projections ($W_Q$, $W_K$, $W_V$, $W_O$) are quantized to
4-bit integers per output channel and dequantized in place. Biases are
unchanged, and activations remain in fp32. The cross-scale replication
(Figure~\ref{fig:quantbars}) uses a symmetric grid with levels $-7$ to
$7$. For each output channel, the step size is chosen from 96
fractions of the max-abs step to minimize the squared weight error.
For each model, this replication quantizes the number of top retrieval
heads whose deficit is closest to 0.5 logits. Pythia-1.4B and 2.8B do
not reach this deficit at 4 bits and use their largest-deficit
settings ($k=40$ and $k=170$). The single-model replication at 410M
(Figure~\ref{fig:quantgap}) sweeps $k$ with a deliberately poor
calibration: an asymmetric min--max grid whose range is stretched
threefold on the side of the larger extreme.

\paragraph{Pruning.}
Pruning zeroes the outputs of half of all attention heads at all
positions. Masks are drawn uniformly at random from a seed, and masks
that would remove a copy-suppression head are rejected. Unlucky masks
are the three seeds, among up to 200{,}000, with the largest overlap
with the top ten retrieval heads; typical masks are the first three
seeds that cover at most five of them. $D$ is the summed clean direct
contribution of the pruned heads, and the accounting is otherwise
unchanged.

\paragraph{Accounting.}
For a target set $T$, the deficit is
$D=\sum_{h\in T}(a_h^{\mathrm{clean}}-a_h^{\mathrm{int}})$ and the
harm is $H=m^{\mathrm{clean}}-m^{\mathrm{int}}$, where $a_h$ is a
direct contribution and $m$ is the score. Release is
$R=\sum_{h\in\mathrm{CS}}(a_h^{\mathrm{int}}-a_h^{\mathrm{clean}})$,
and standing suppression is
$S=-\sum_{h\in\mathrm{CS}}a_h^{\mathrm{clean}}$. The backup response
$B$ is defined in the same way over all other heads and MLP layers.
The normalization term $N$ is the change in the part of $m$ that the
component contributions, read at the clean scale, do not account for.
With these definitions, Equation~\ref{eq:accounting} holds up to
the small numerical residuals reported with the decompositions.
Fractional repair is $(\bar{D}-\bar{H})/\bar{D}$, a ratio of means
over prompts.

\paragraph{Training checkpoints and logit lens.}
The checkpoint analysis uses saved checkpoints of Pythia-410M, 1B,
1.4B, and 6.9B at steps 1k, 5k, 10k, 20k, 40k, 70k, 100k, and 143k,
with two further steps per model selected on discovery measurements.
Intermediate checkpoints of 2.8B and 12B are excluded because their
source identities could not be resolved. The logit lens applies the
final layer normalization to the residual stream after each layer,
with the normalization scale recomputed at that layer
\citep{nostalgebraist2020logitlens,belrose2023tuned}.
Appendix~\ref{sec:app-training-codevelopment} gives the checkpoint
procedure, and Appendix~\ref{sec:app-readout-sensitivity} compares
readout conventions.

\paragraph{Statistics.}
Intervals in the held-out analyses are 95\% paired template-cluster
bootstrap intervals with 2{,}000 replicates: prompts are resampled by
template, and the same prompts are shared across models and
conditions. These intervals describe variation over prompts, not over
independently trained models. The exploratory analyses report means
over the 1{,}000 prompts.

Table~\ref{tab:models} lists clean-run statistics and head-set sizes.
The number of identified retrieval heads varies across models.
Retrieval-head ablations target the highest-ranked heads in each
model; the deficit $D$ measures the contribution removed by each
intervention.

\begin{table}
  \centering
  \begin{tabular}{lrrrr}
    \hline
    \textbf{Model} & \textbf{Acc.} & \textbf{Logit diff} &
    $|R|$ & $|CS|$ \\
    \hline
    410M & 0.99 & 4.31 & 98  & 2 \\
    1B   & 1.00 & 5.05 & 27  & 2 \\
    1.4B & 1.00 & 4.21 & 105 & 2 \\
    2.8B & 1.00 & 4.30 & 283 & 2 \\
    6.9B & 1.00 & 3.95 & 266 & 2 \\
    12B  & 1.00 & 4.32 & 389 & 2 \\
    \hline
  \end{tabular}
  \caption{Clean-run statistics per model on the exploratory prompts:
  pairwise IOI accuracy, clean logit difference, and the sizes of the
  retrieval ($R$) and reference copy-suppression ($CS$) sets.}
  \label{tab:models}
\end{table}

\section{Repair Mechanisms}
\label{sec:appendix-detail}

These analyses extend the decomposition in Section~\ref{sec:single}
to additional models and examine how individual response channels
depend on the intervention.

\subsection{Decompositions across Model Families}

\begin{figure*}[t]
  \centering
  \includegraphics[width=0.48\textwidth]{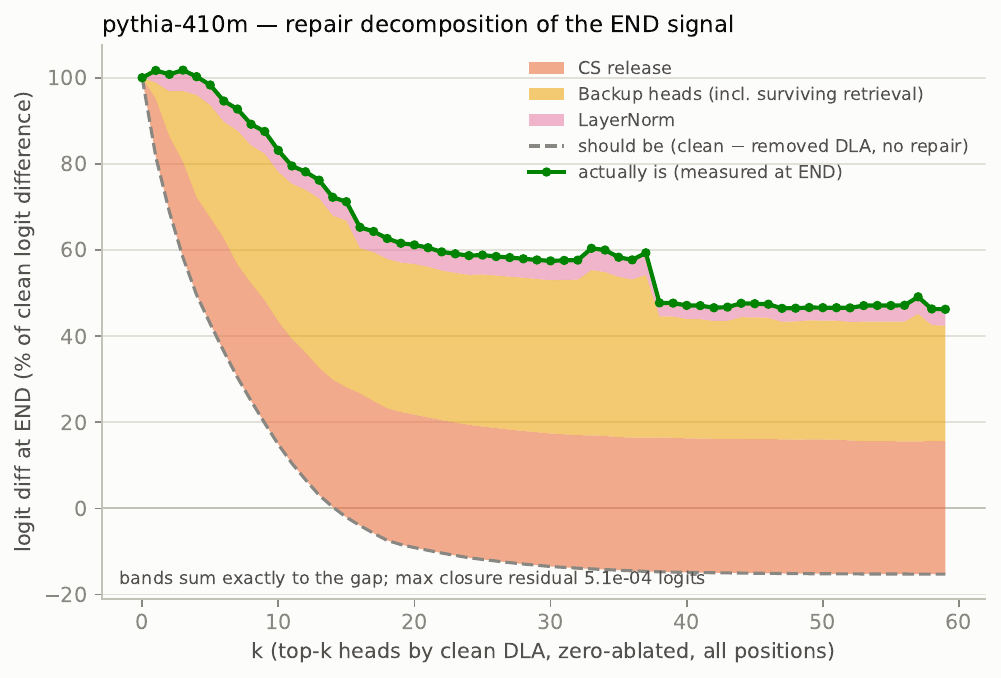}\hfill
  \includegraphics[width=0.48\textwidth]{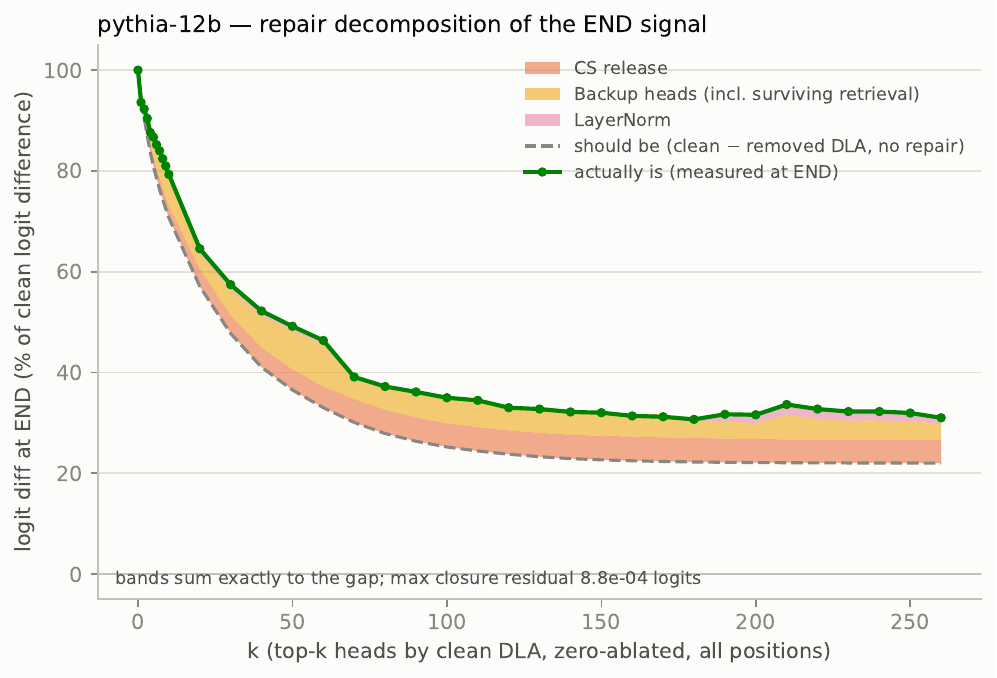}
  \par\medskip
  \includegraphics[width=0.48\textwidth]{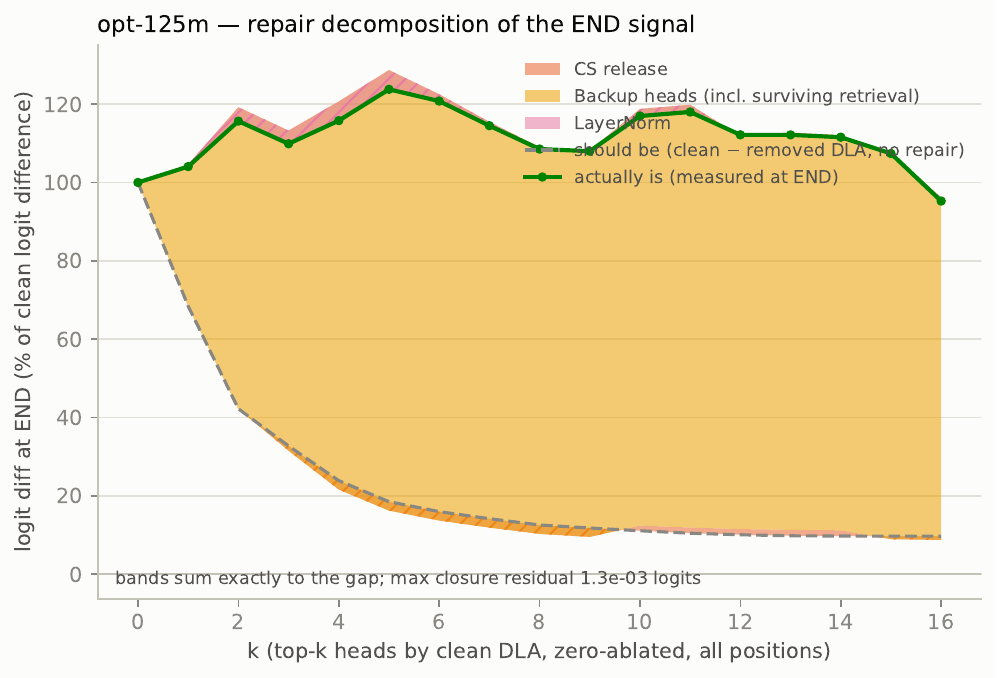}\hfill
  \includegraphics[width=0.48\textwidth]{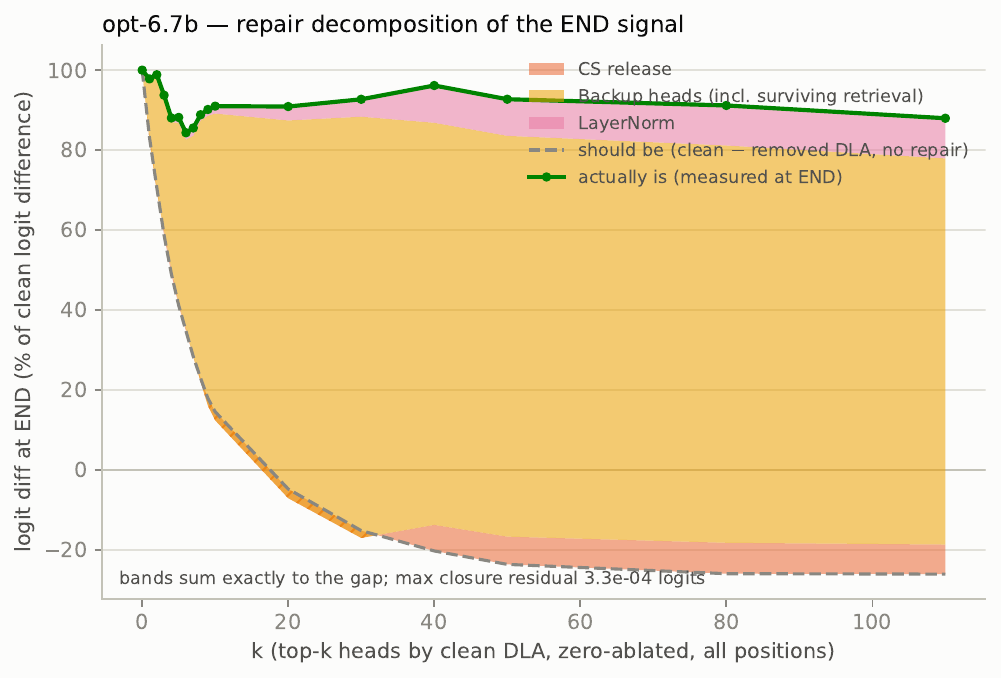}
  \caption{Repair decomposition under retrieval-head ablation in
  Pythia-410M and Pythia-12B (top row), and OPT-125M and OPT-6.7B
  (bottom row). The smaller model is on the left in each row.
  The dashed curve is the no-response baseline,
  $m^{\mathrm{clean}}-D$; the solid curve is the measured output
  margin. Their difference is total repair, decomposed into
  copy-suppression release, the backup response, and final
  normalization. Values are percentages of each model's clean
  logit difference. Both OPT models retain substantial repair
  across the displayed sweeps. The panels have different ranges
  of $k$ and do not match damage across models.}
  \label{fig:app-repair-bands}
\end{figure*}

\begin{figure}[t]
  \centering
  \includegraphics[width=\columnwidth]{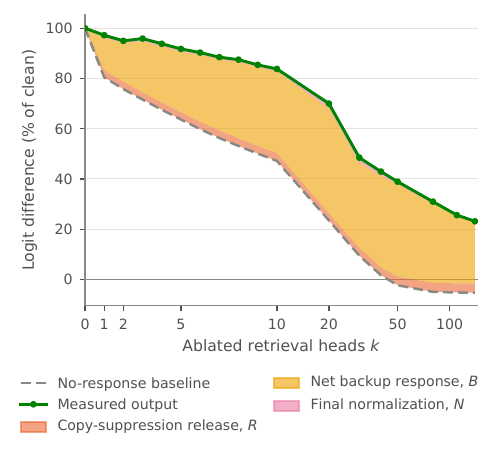}
  \caption{Repair decomposition in Llama 3.1 8B as the top-$k$
  retrieval heads are ablated. The gap between the no-response
  baseline and the measured output is filled by copy-suppression
  release, the net backup response, and final normalization.
  Values are percentages of the clean logit difference.}
  \label{fig:llama-bands}
\end{figure}

Figure~\ref{fig:app-repair-bands} compares the repair decomposition in
Pythia and OPT, while Figure~\ref{fig:llama-bands} gives a separate
Llama replication. The gap between the measured output and the
no-response baseline identifies repair in each sweep; the sweeps
do not match the removed contribution across models.

\subsection{Component Responses and Path Cuts}

\begin{figure}[t]
  \centering
  \includegraphics[width=\columnwidth]{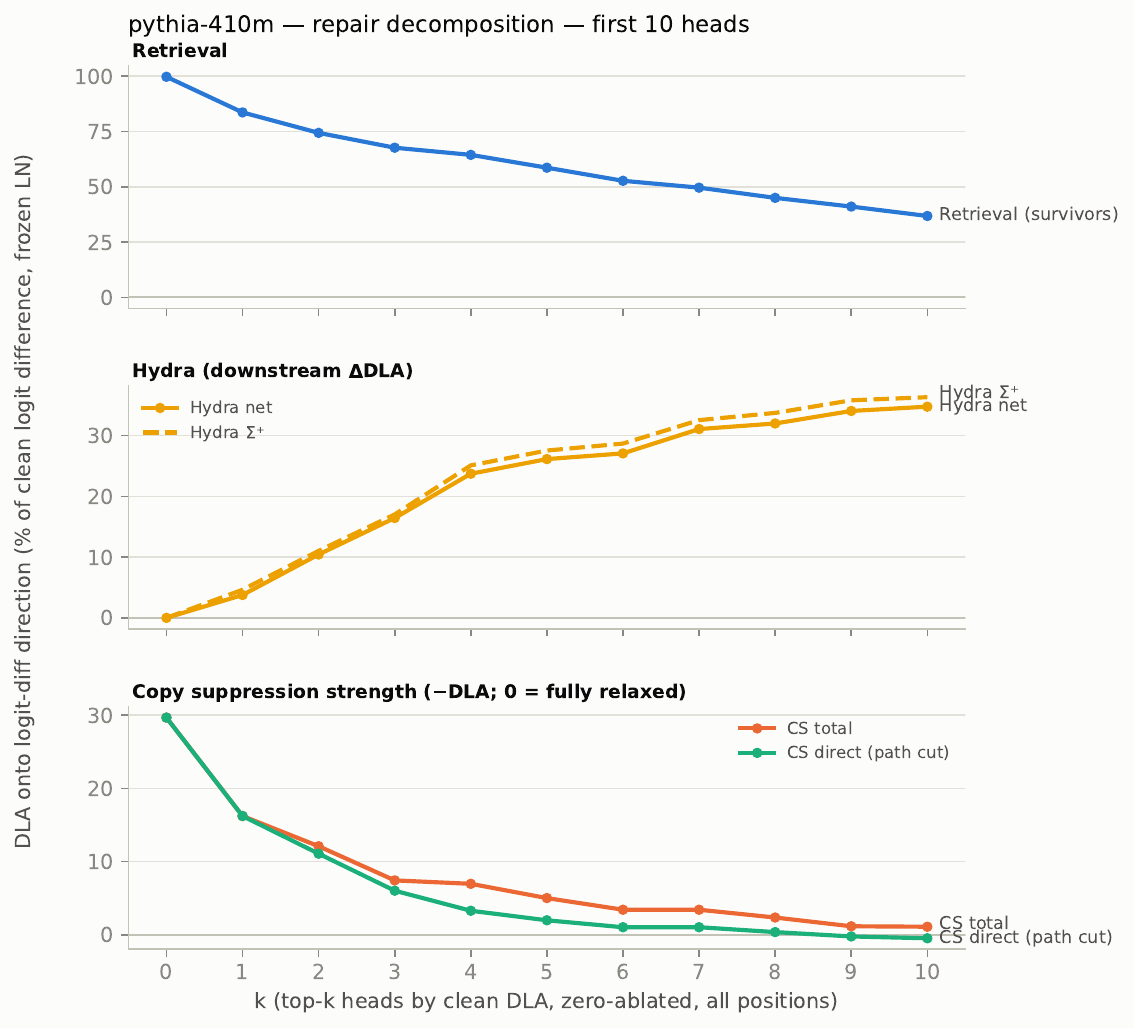}
  \caption{Response to the first ten retrieval-head ablations in
  Pythia-410M. Top: surviving retrieval contribution. Middle: net
  and summed positive backup responses. Bottom: copy-suppression
  strength after full ablation and a path cut; zero indicates
  complete release. Values are percentages of the clean logit
  difference, using the clean final-normalization scale.}
  \label{fig:c2channels}
\end{figure}

Two finer-grained views support the channel analysis of
Section~\ref{sec:mechanism}. First, Figure~\ref{fig:c2channels}
tracks the Pythia-410M response as the first ten retrieval heads
are removed one at a time. The surviving retrieval contribution
declines, the net backup response increases, and copy-suppression
strength falls toward zero. By the tenth head, the standing
suppression is almost fully released.

\begin{figure}[t]
  \centering
  \includegraphics[width=\columnwidth]{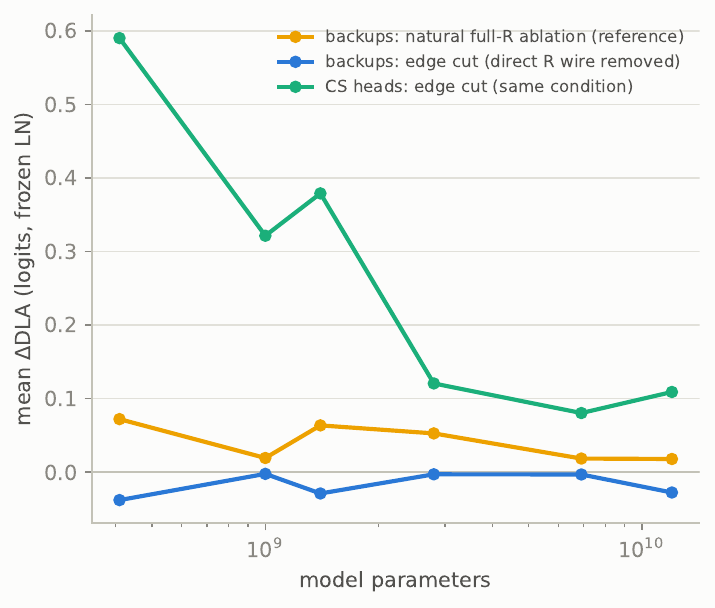}
  \caption{Mean change in direct contribution per selected head
  across Pythia, in logits under clean final normalization.
  Backup-head responses are positive under full retrieval-head
  ablation and near-zero or slightly negative when their direct
  retrieval input is removed (edge cut). Copy-suppression heads
  increase their contribution under the cut. Removing the direct
  input alone therefore does not reproduce the backup response
  to full ablation.}
  \label{fig:c3edge}
\end{figure}

Second, Figure~\ref{fig:c3edge} compares full retrieval-head ablation
with removing their direct contribution only from selected
downstream head inputs. The selected backup heads show a positive
mean change in direct contribution after full ablation and a
near-zero or slightly negative change after the path cut. Copy-suppression heads instead
increase their direct contribution under the path cut. Thus, removing
the direct input alone does not reproduce the backup response to
full ablation. This contrast is consistent with a response to changes
in the wider computation, while copy-suppression is sensitive to
the direct retrieval input.

\section{Scaling Comparisons}
\label{sec:app-scaling}

These comparisons examine how repair changes with model size
under different intervention designs. We first examine a shared deficit in OPT,
then calibrated held-out ablations and response prevention, and
finally the exploratory comparisons that fix the number of heads.

\subsection{Shared-Deficit Comparison in OPT}
\label{sec:app-opt-scaling}

\begin{figure}[t]
  \centering
  \includegraphics[width=\columnwidth]{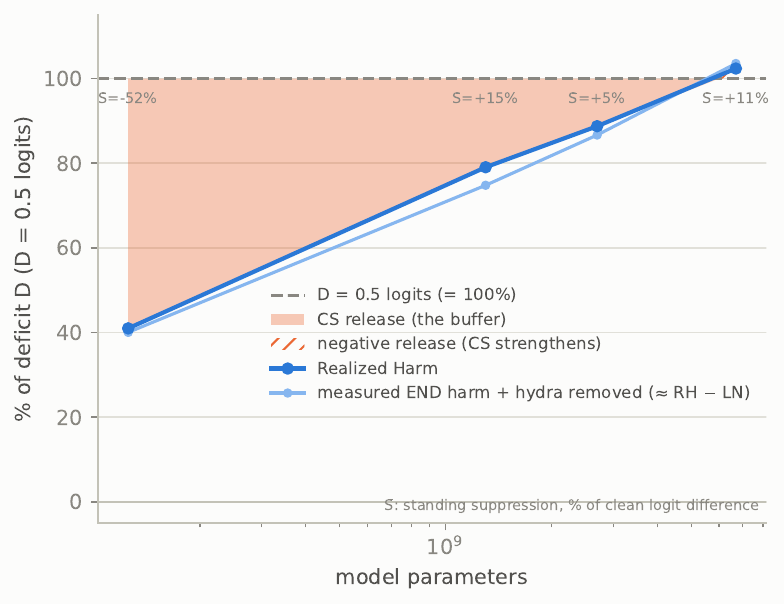}
  \caption{Deficit remaining after copy-suppression release at
  $D=0.5$ logits in OPT-125M, 1.3B, 2.7B, and 6.7B. The dashed
  line marks the initial deficit; the dark blue curve shows the
  remainder $D-R$. Shading indicates release, with a remainder
  above the dashed line indicating negative release. The light
  blue curve shows output harm plus the backup response,
  $H+B=D-R-N$. Both curves are percentages of $D$; neither is
  final output harm alone. The $S$ annotations give standing
  suppression as a percentage of each model's clean logit
  difference.}
  \label{fig:app-opt-shared-deficit}
\end{figure}

Figure~\ref{fig:app-opt-shared-deficit} extends the shared-deficit
comparison of Figure~\ref{fig:c3harm} to OPT. At $D=0.5$ logits,
the fraction remaining after copy-suppression release increases
across the four model sizes and slightly exceeds the initial
deficit at 6.7B. This comparison measures the remainder $D-R$;
final output harm also depends on the backup response and final
normalization.

\subsection{Comparable Damage and Natural Responses}
\label{sec:app-backup-evaluation}

To compare response channels under similar retrieval-head damage
and measure the effect of preventing selected responses, we evaluate
all six final Pythia models on 2{,}000 fresh, deduplicated held-out IOI
prompts shared across model sizes.
Target ranking, intervention strength, and receiver selection
are determined on discovery prompts and frozen before held-out
evaluation. These experiments are separate from the original
1{,}000-prompt ablation sweeps and their interpolated
$D=0.5$ comparison.

\begin{figure*}[t]
  \centering
  \includegraphics[width=0.66\textwidth]{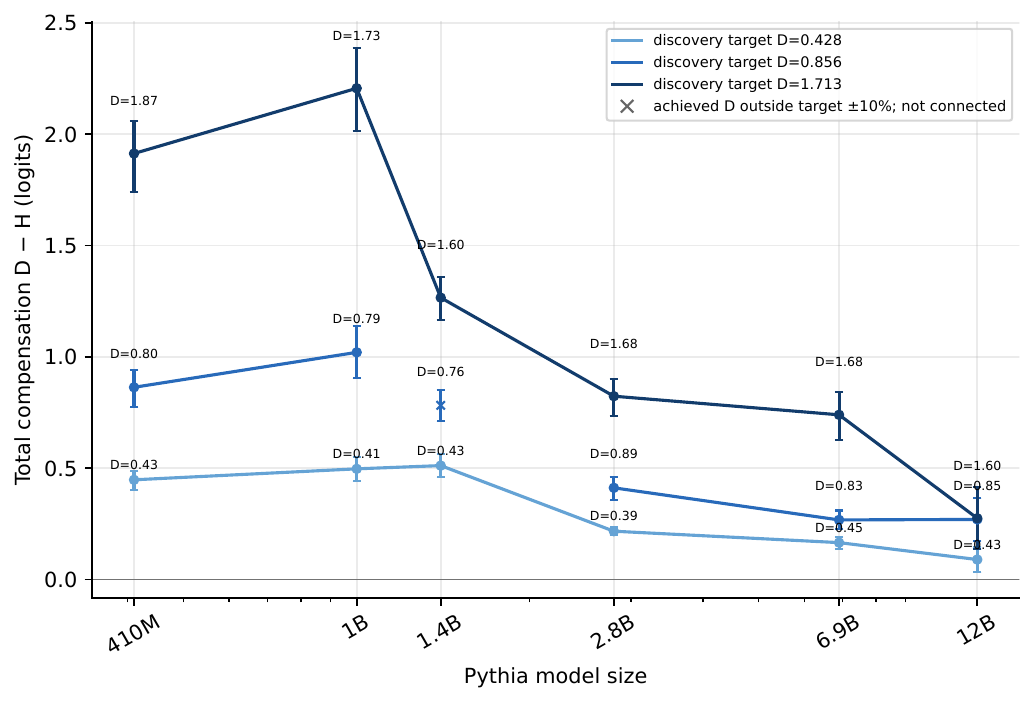}
  \caption{Total repair at three discovery-calibrated ablation
  strengths. Each point reports the achieved held-out deficit
  $D$; error bars are 95\% paired template-cluster bootstrap
  intervals for total repair $D-H$. Crosses mark conditions
  outside the $\pm10\%$ tolerance around the discovery target
  and are not connected as a matched-damage series.
  The middle target supplies the primary response-channel
  and response-prevention comparisons in the main text.}
  \label{fig:app-matched-damage-ablation}
\end{figure*}

We use contribution-ranked whole-head zeroing when it reaches
a discovery target within 10\%. Otherwise, we attenuate the
boundary head's live output to approach the target.
The attenuated output is computed from the current intervened
state; it is not replayed from a clean run.
For this evaluation, $D$ is the measured difference between
the targets' clean and intervened direct contributions under
the clean final-normalization readout.
The three discovery targets are approximately 0.428, 0.856,
and 1.713 logits. Figure~\ref{fig:app-matched-damage-ablation}
reports achieved held-out damage and total repair at each
target. The middle target supplies the primary conditions
in Figures~\ref{fig:c3response} and~\ref{fig:c3protect}.
Its achieved deficits range from 0.756 to 0.888 logits;
the 1.4B condition falls outside the frozen $\pm10\%$
tolerance. The held-out conditions are not retuned.

\begin{figure*}[t]
  \centering
  \includegraphics[width=0.90\textwidth]{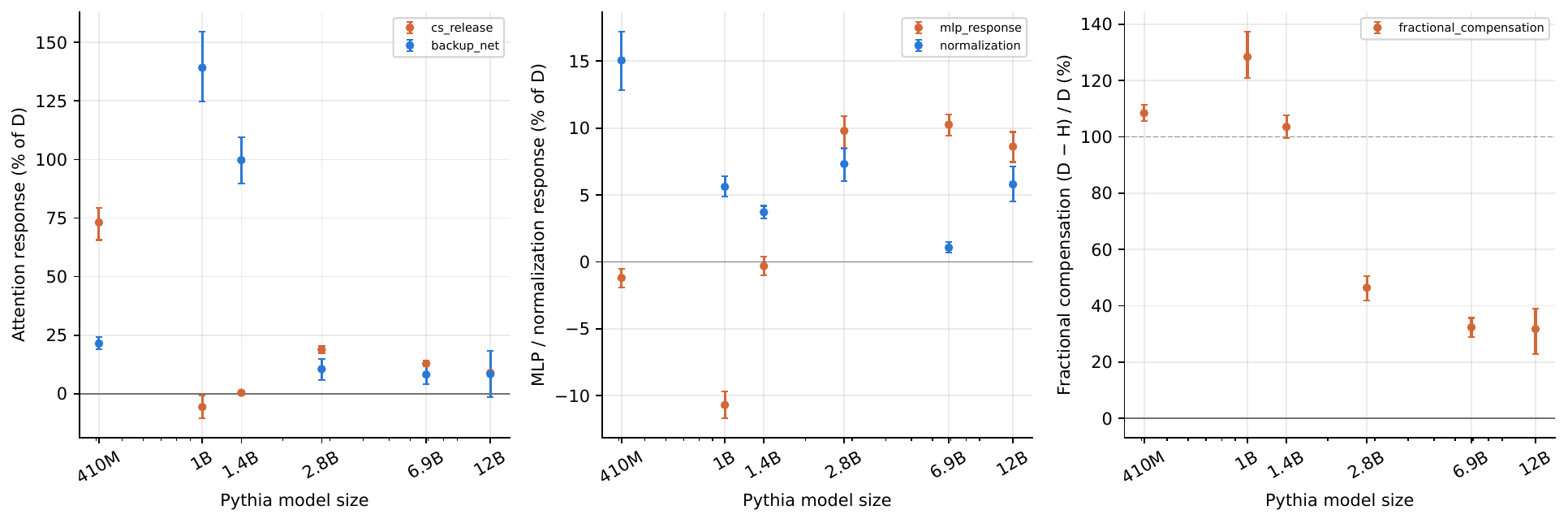}
  \caption{Full response-channel decomposition at the primary
  ablation condition used in Figure~\ref{fig:c3response}.
  Left: copy-suppression release and the net attention response.
  Centre: MLP and final normalization responses. Right:
  fractional repair, $(D-H)/D$. All values are percentages of
  the deficit $D$.
  Error bars are 95\% paired template-cluster bootstrap intervals
  over the held-out prompts. The attention and MLP responses
  together form the broader backup term $B$.
  Achieved deficits, including the 1.4B condition outside the
  discovery tolerance, are reported in
  Table~\ref{tab:app-backup-response}.}
  \label{fig:app-response-channels}
\end{figure*}

Figure~\ref{fig:app-response-channels} gives the full decomposition
into CS release, the attention response, the MLP response, and
normalization, together with total fractional repair.
Table~\ref{tab:app-backup-response}(a) reports the corresponding
means at the primary level. The attention and MLP contributions
together form the backup response, $B=A+M$.

\subsection{Response Prevention and Controls}
\label{sec:app-response-prevention}

For response prevention, backup receivers are selected as the
smallest eligible set covering 90\% of the positive attention
response on discovery prompts. Receivers exclude the primary
targets and copy-suppression heads. We replace their output
vectors before the output projection with their clean-run values
at the answer position. Their clean contribution is preserved,
while subsequent computation remains live.
Separate conditions hold the paper-reference copy-suppression
heads clean, the backup receivers clean, or both populations
clean. Low-response heads and clean-to-clean replacement provide
controls; an additional functional-CS comparison is retained
where available.

\begin{figure*}[t]
  \centering
  \includegraphics[width=0.72\textwidth]{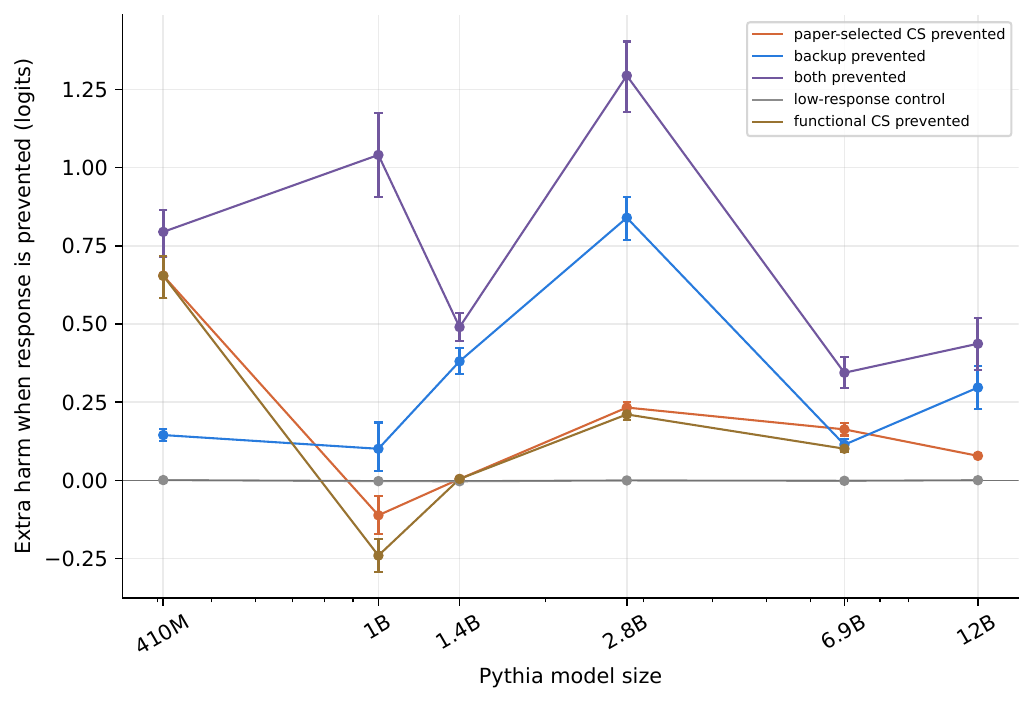}
  \caption{Extra output harm when selected responses to retrieval-head
  damage are prevented. Ablations produce deficits of about 0.8 logits
  on 2{,}000 held-out prompts, as in Figure~\ref{fig:c3response}.
  Selected heads are held at their clean output at the answer
  position while subsequent computation remains live. Extra harm
  is the increase in $H$ relative to the natural response.
  Backup receivers are selected to cover 90\% of the positive
  attention response on discovery prompts; low-response heads
  provide a control. The functional-CS curve
  uses a separate head selection from the paper-reference CS set.
  Error bars are 95\% paired template-cluster bootstrap intervals.
  The 1.4B backup and joint clamps also change the removed
  contribution by about $-0.05$ logits; backup receiver counts and
  condition-specific deficits are given in
  Appendix~Table~\ref{tab:app-backup-response}.}
  \label{fig:c3protect}
\end{figure*}

Figure~\ref{fig:c3protect} tests whether the selected responses
protect the prediction by measuring the extra harm when their
adjustment is prevented. Low-response heads provide a control for
applying the same intervention to a different head population.

Table~\ref{tab:app-backup-response} distinguishes the signed
attention contribution from the effect of preventing the selected
response. The signed sum includes all surviving non-CS heads,
including surviving retrieval heads; the selected receiver count
describes only the positive-response set used by the clamp.
Extra harm includes the effects of subsequent computation and
need not equal that signed direct contribution.
We measure $D$ in every clamp condition: the 1.4B backup and
joint clamps change it by approximately $-0.049$ logits because
the attenuated boundary head responds to upstream changes.

Intervals use 2{,}000 paired template-cluster bootstrap
replicates and describe prompt variation, rather than variation
across independently trained model seeds.

\begin{table*}
  \centering
  \small
  \setlength{\tabcolsep}{5pt}
  \textbf{(a) Natural response at the primary ablation condition}\\[3pt]
  \begin{tabular}{lrrrrrrr}
    \toprule
    Model & $D$ & $R$ & $A$ & $M$ & $N$ & $H$ & $D-R$ \\
    \midrule
    410M & 0.797 & 0.583 & 0.171 & $-0.010$ & 0.120 & $-0.067$ & 0.214 \\
    1B   & 0.795 & $-0.045$ & 1.106 & $-0.085$ & 0.045 & $-0.226$ & 0.840 \\
    1.4B & 0.756 & 0.003 & 0.754 & $-0.002$ & 0.028 & $-0.027$ & 0.753 \\
    2.8B & 0.888 & 0.167 & 0.093 & 0.087 & 0.065 & 0.476 & 0.721 \\
    6.9B & 0.829 & 0.106 & 0.068 & 0.085 & 0.009 & 0.561 & 0.722 \\
    12B  & 0.851 & 0.076 & 0.071 & 0.073 & 0.049 & 0.581 & 0.775 \\
    \bottomrule
  \end{tabular}

  \medskip
  \textbf{(b) Response prevention}\\[3pt]
  \begin{tabular}{lrcrrr}
    \toprule
    Model & $n$ & $\Delta H_{\mathrm{backup}}$ [95\% CI] &
    $H_{\mathrm{both}}$ & $D_{\mathrm{both}}$ &
    $\Delta D_{\mathrm{backup}}$ \\
    \midrule
    410M & 9  & 0.145 [0.127, 0.164] & 0.728 & 0.797 & 0.000 \\
    1B   & 1  & 0.101 [0.029, 0.185] & 0.815 & 0.795 & 0.000 \\
    1.4B & 14 & 0.381 [0.341, 0.422] & 0.464 & 0.707 & $-0.049$ \\
    2.8B & 19 & 0.840 [0.768, 0.905] & 1.770 & 0.888 & 0.000 \\
    6.9B & 32 & 0.114 [0.095, 0.134] & 0.905 & 0.829 & 0.000 \\
    12B  & 34 & 0.297 [0.228, 0.366] & 1.018 & 0.851 & 0.000 \\
    \bottomrule
  \end{tabular}
  \caption{Natural response and the effect of response prevention
  at the primary ablation condition, in logits. In (a), $A$ is
  the signed non-CS attention response and $M$ is the MLP
  response, so the manuscript's broader backup term is $B=A+M$.
  All entries are means. In (b), $n$ is the number of selected
  backup receivers and $\Delta H_{\mathrm{backup}}$ is extra
  harm when their adjustment is prevented, with 95\% paired
  template-cluster bootstrap intervals. The joint condition
  prevents both backup and paper-reference CS adjustment;
  $H_{\mathrm{both}}$ and $D_{\mathrm{both}}$ are its measured
  harm and deficit. The final column records the change in
  $D$ under the backup clamp; the same drift occurs under the
  joint clamp. The attention response and the causal effect
  concern different receiver populations and measurements.}
  \label{tab:app-backup-response}
\end{table*}

\begin{figure*}[t]
  \centering
  \includegraphics[width=0.90\textwidth]{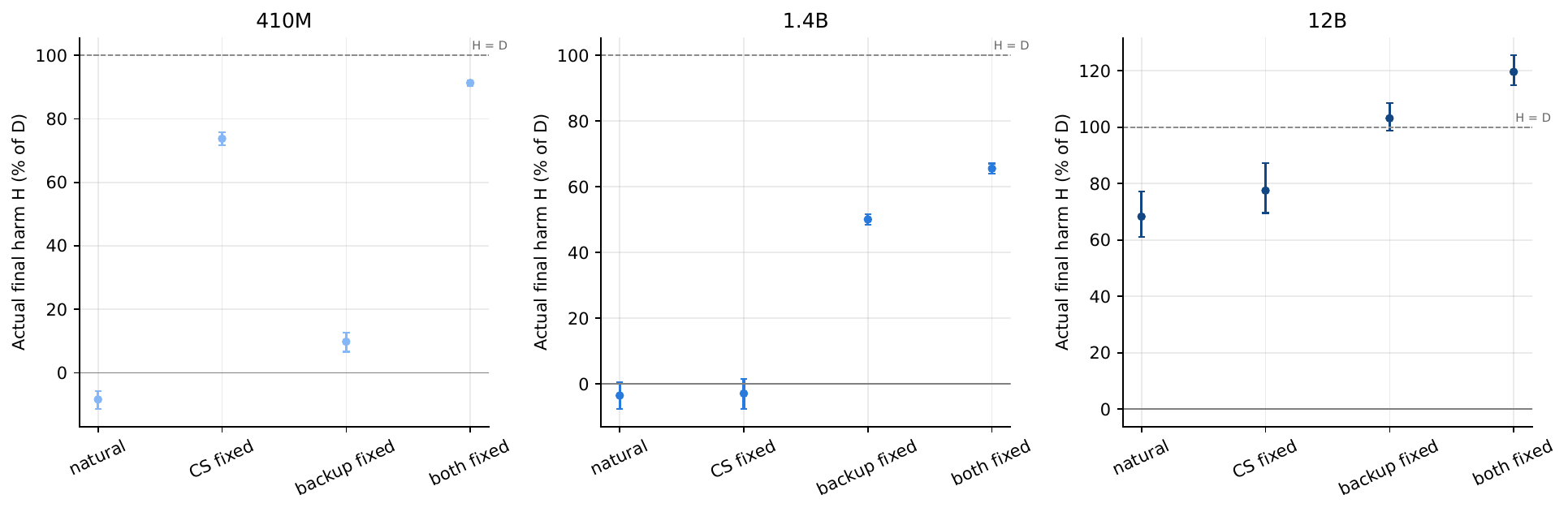}
  \caption{Absolute output harm under natural response and
  response prevention at 410M, 1.4B, and 12B. Panels show the
  natural response, CS held clean, backup receivers held clean,
  and both held clean. Harm is shown as a percentage of the
  deficit $D$; the dashed line marks $H=D$. Natural deficits are approximately
  0.80, 0.76, and 0.85 logits, respectively. At 1.4B, the
  backup and joint clamps reduce the measured deficit to
  about 0.71 logits. Error bars are 95\% paired template-cluster
  bootstrap intervals. Table~\ref{tab:backup-harm}
  gives natural, backup, and joint-condition harm across all six models.}
  \label{fig:app-harm-conditions}
\end{figure*}

Figure~\ref{fig:app-harm-conditions} provides the corresponding
absolute-harm view for three representative models.
This view places the natural response and prevention conditions
relative to the no-repair reference, $H=D$.

\subsection{Fixed Ten-Head Comparisons}
\label{sec:app-ten-head}

\begin{figure}[t]
  \centering
  \includegraphics[width=\columnwidth]{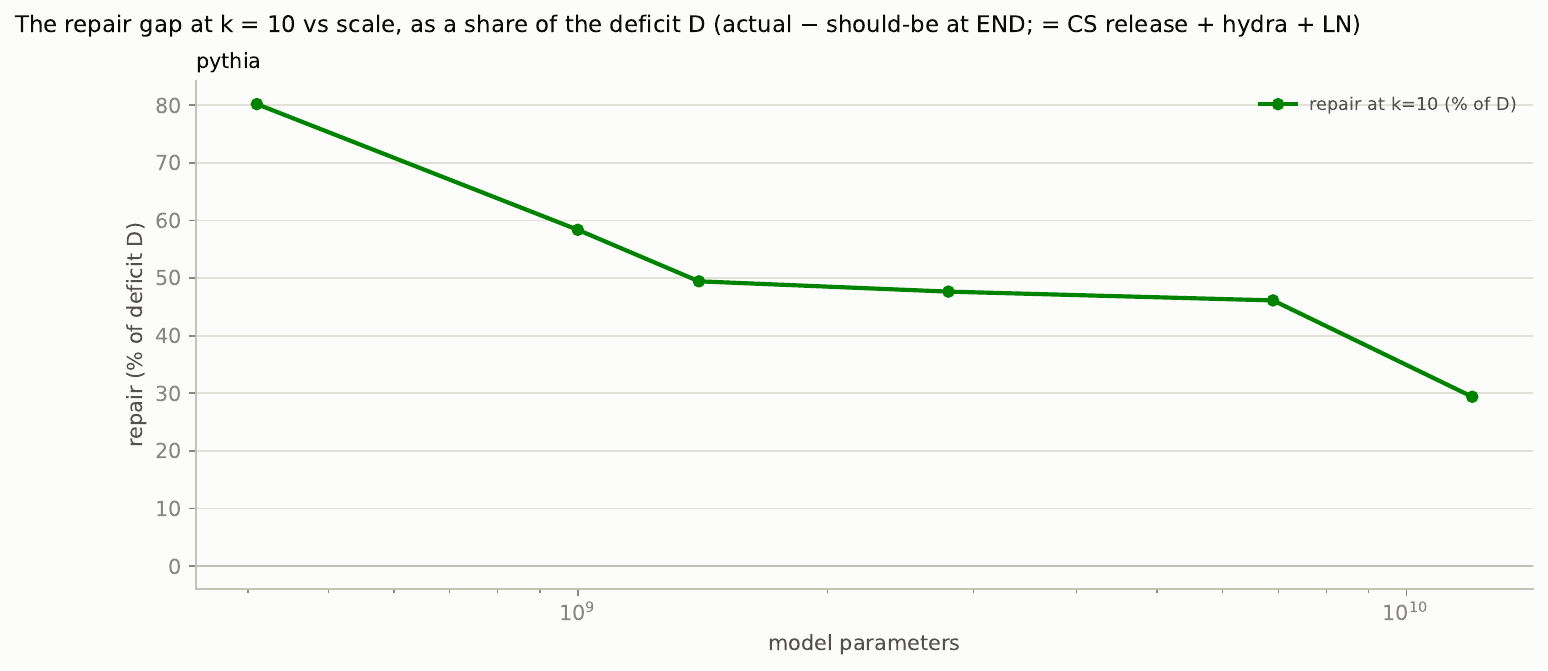}
  \caption{Total repair in the original ten-head ablation
  comparison: copy-suppression release, backup response, and
  normalization summed. Repair is shown as a percentage of each
  model's deficit, which differs across model sizes. It falls
  from about 80\% at 410M to about 29\% at 12B in this
  exploratory evaluation.}
  \label{fig:repairk10}
\end{figure}

\begin{figure}[t]
  \centering
  \includegraphics[width=\columnwidth]{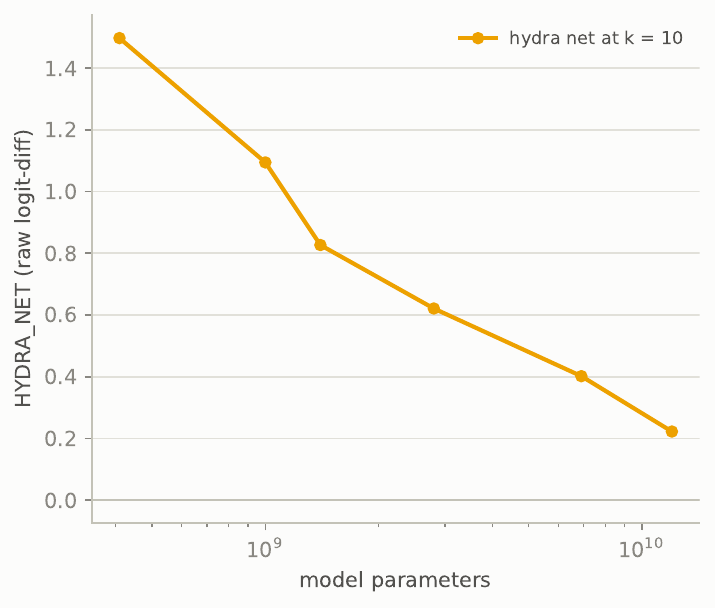}
  \caption{Backup response in the original ten-head ablation
  comparison on exploratory prompts. Values are in logits. The
  removed contribution varies from approximately 1.3 to 3.7 logits
  across models.
  This fixed-head-count view is separate from the held-out
  comparable-damage evaluation in Figure~\ref{fig:c3response}.}
  \label{fig:c3hydra}
\end{figure}

Figures~\ref{fig:repairk10} and~\ref{fig:c3hydra} retain the
original ten-head comparisons on the exploratory prompt set.
They fix the number of ablated heads, while the removed direct
contribution varies from approximately 1.3 to 3.7 logits across
models. They therefore describe a different comparison from
the shared-deficit interpolation and the held-out calibrated
ablations in Section~\ref{sec:scale}. These original measurements
are not pooled with the newer held-out results.

\section{Compression Replications}
\label{sec:app-compression}

These experiments test whether the damage--repair distinction also
appears under weight quantization and structural pruning. Their
intervention scopes differ: quantization targets retrieval heads,
whereas pruning can also remove components that contribute to repair.

\subsection{Quantization}
\label{sec:appendix}

The main analysis uses ablation because it makes the removed quantity
exact; here we repeat the accounting with quantization as the damage.
The targeted retrieval heads' four attention projections are quantized
per output channel to 4-bit integers (Section~\ref{sec:method}). The
deficit is the resulting drop in those heads' direct contributions.
For the cross-scale comparison, the number of heads is selected to
give a deficit closest to 0.5 logits, with the calibration and
fallback settings described in Appendix~\ref{sec:app-setup}.

\begin{figure}[t]
  \centering
  \includegraphics[width=\columnwidth]{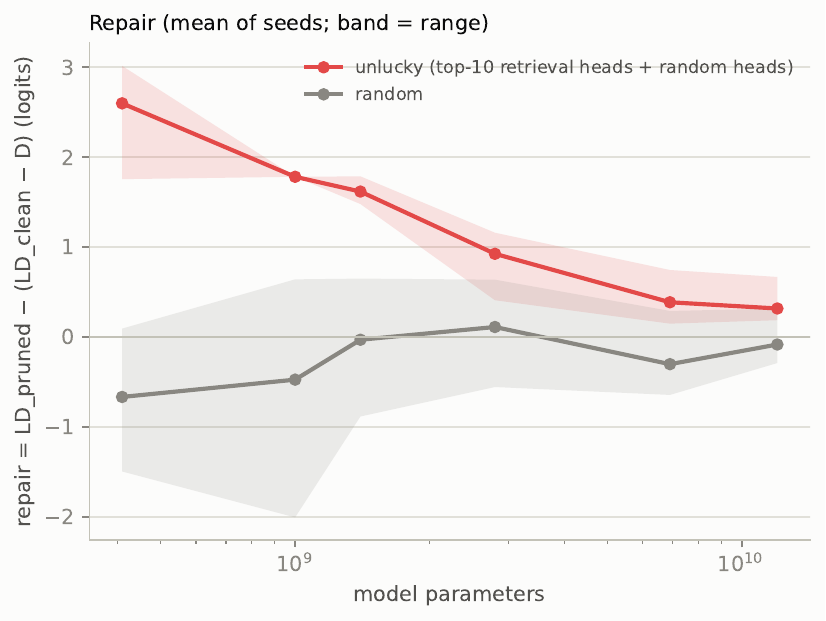}
  \caption{Repair under structural pruning of half the eligible
  heads across the Pythia suite. Red shows unlucky masks with high
  overlap with the top ten retrieval heads; gray shows random
  comparison masks. Lines are means over three seeds, and shaded
  bands span their ranges. Repair is $D-H$, the measured output
  minus the no-response baseline; negative values indicate harm
  beyond the removed direct contribution. Unlucky-mask repair
  decreases while remaining positive, whereas random-mask repair
  is non-monotonic.}
  \label{fig:prunerepair}
\end{figure}

Figure~\ref{fig:quantgap} shows that Pythia-410M retains a gap between
the no-response baseline and the measured output under quantization.
That single-model sweep uses a different calibration from the scaling
comparison, as described in Appendix~\ref{sec:app-setup}.
Figure~\ref{fig:quantbars} compares total output harm relative to the
achieved deficit, $H/D$, in Pythia and OPT. In Pythia, this fraction
rises overall from 0.55 at 410M to 0.91 at 12B, with variation at
intermediate sizes. In OPT, it rises from 0.41 at 125M to 0.98 at
6.7B. Thus, the quantization comparison shows a greater share of the
deficit retained as output harm in the larger models of both suites.

\begin{figure*}[t]
  \centering
  \begin{minipage}[t]{0.48\textwidth}
    \centering
    \includegraphics[width=\linewidth]{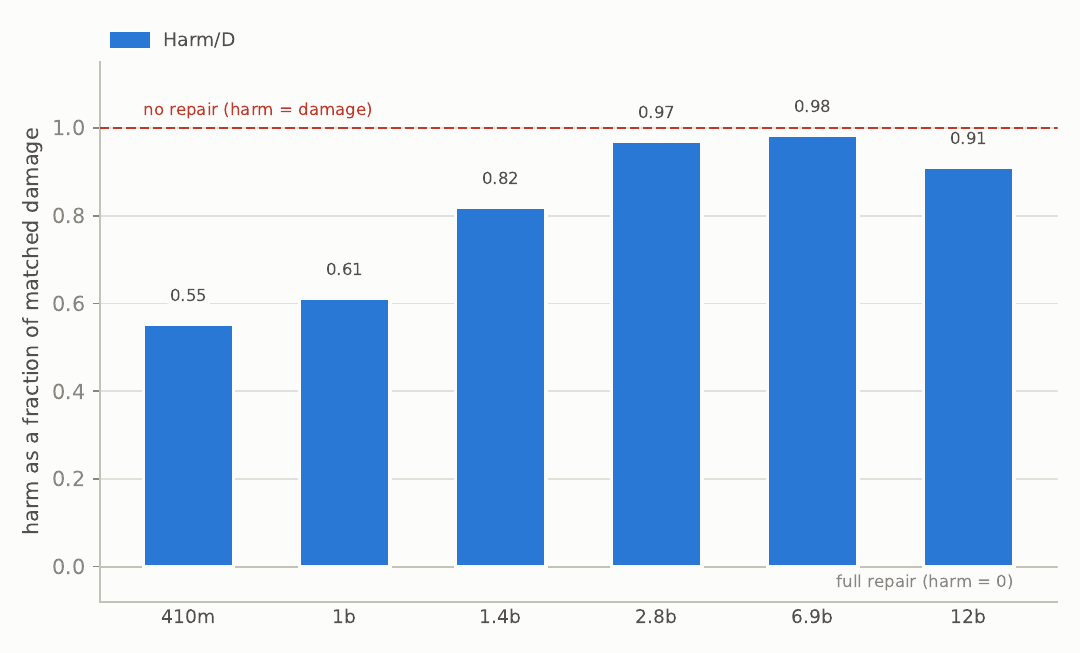}
  \end{minipage}\hfill
  \begin{minipage}[t]{0.48\textwidth}
    \centering
    \includegraphics[width=\linewidth]{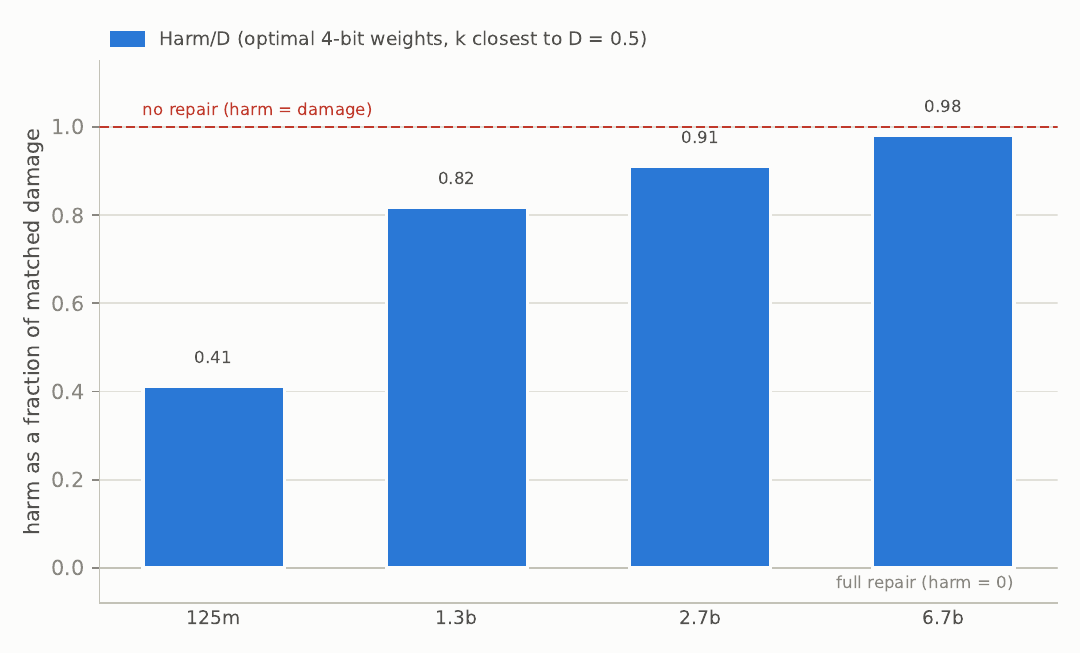}
  \end{minipage}
  \caption{Output harm as a fraction of the achieved deficit under
  4-bit retrieval-head weight quantization in Pythia (left) and
  OPT (right). For each model, the number of quantized heads is
  chosen to give a deficit closest to 0.5 logits; Pythia-1.4B and
  2.8B use their largest-deficit settings because they do not
  reach this target. Bars and labels show $H/D$, and the dashed
  line marks no net repair, $H=D$. The selected settings need not
  have identical deficits.}
  \label{fig:quantbars}
\end{figure*}

\subsection{Pruning}
\label{sec:appendix-prune}

We prune half of the eligible heads with structural masks under two
conditions: unlucky masks with high overlap with the top ten
retrieval heads and random comparison masks (three seeds each;
Appendix~\ref{sec:app-setup}). Unlike targeted retrieval-head
ablation, these masks can also remove components that contribute
to repair.

Figure~\ref{fig:prunerepair} plots total repair, the measured output
minus the no-response baseline. Under unlucky masks, mean repair
remains positive but decreases with model size. Under random masks,
mean repair varies around zero without a monotonic scale trend and
is lower than under unlucky masks at each displayed size. The
reduction in repair with scale is therefore clearest for the
unlucky condition.

\section{Training Dynamics}
\label{sec:app-training-codevelopment}

\begin{figure*}[t]
  \centering
  \includegraphics[width=0.90\textwidth]{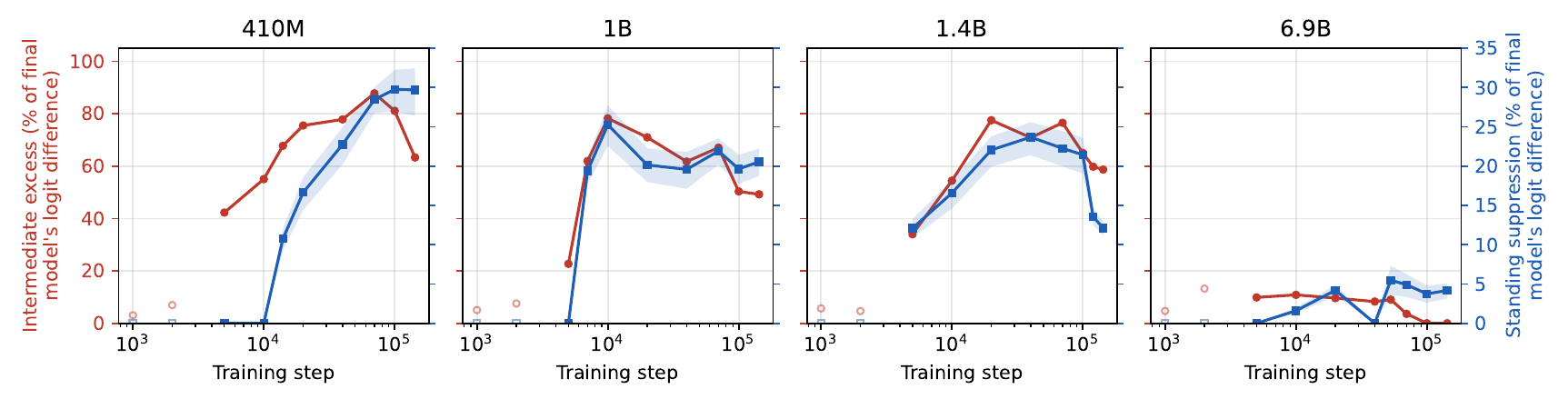}
  \caption{Co-development of intermediate excess and standing
  suppression on clean runs across saved training checkpoints.
  Red circles, left axis: the maximum of the mean
  answer--distractor logit-difference trajectory minus that
  checkpoint's own final-layer value, using the layer-renormalized
  logit lens. Blue squares, right
  axis: standing suppression of the copy-suppression heads
  identified at that checkpoint. Both quantities are percentages
  of the final model's clean logit difference. Shading gives 95\% paired
  template-cluster bootstrap intervals for suppression; the
  excess is a mean-curve statistic without an interval.
  Hollow markers indicate pairwise IOI accuracy below 0.6.
  Axis scales are shared across models but differ between
  the two quantities. Only the four models with usable
  intermediate checkpoint histories are shown.}
  \label{fig:app-training-codevelopment}
\end{figure*}

Figure~\ref{fig:app-training-codevelopment} examines whether
intermediate excess and standing suppression develop together
during training. This within-model comparison complements the
final-model scaling results; it measures association rather than
the effect of an intervention on training.

To examine how intermediate excess and standing suppression develop
within each model, we evaluate the selected training checkpoints of
Pythia-410M, 1B, 1.4B, and 6.9B on the same 2{,}000 held-out IOI
prompts.
The initial checkpoint grid is 1k, 5k, 10k, 20k, 40k, 70k,
100k, and 143k training steps. Two additional checkpoints per
model were selected from discovery measurements around the
largest changes in task competence, intermediate downward
variation, and standing suppression. These additions are 2k
and 14k for 410M, 2k and 7k for 1B, 2k and 120k for 1.4B,
and 2k and 53k for 6.9B. The selection was frozen before
examining the held-out results. Intermediate checkpoints for
2.8B and 12B are excluded because their source identities
could not be resolved; they do not supply training trajectories.

For each checkpoint, we project the residual after each layer
onto the answer--distractor direction using the layer-renormalized
logit lens: each intermediate residual is normalized separately.
We average the resulting logit differences across prompts before
measuring the intermediate excess. If $\bar{s}_{\ell}$ denotes
this mean at layer $\ell$ and $L$ is the final layer, the excess
is $\max_{\ell}\bar{s}_{\ell}-\bar{s}_{L}$.
This is the excess of the mean trajectory, rather than the mean
of the excesses computed separately for individual prompts.

We identify copy-suppression heads independently at each checkpoint
using negative direct contribution, attention to the answer, and
the OV and QK gates described in Section~\ref{sec:method}.
No minimum number of heads is imposed. Standing suppression is
the negative sum of their clean direct contributions, measured
using each prompt's clean final-normalization scale.
This functional selection can yield a different population from
the fixed paper-reference set used in the original final-model
comparison.

Figure~\ref{fig:app-training-codevelopment} compares the two
measurements on clean runs only. Filled markers identify
checkpoints with pairwise IOI accuracy of at least 0.6; earlier,
less competent checkpoints remain visible as hollow markers.
This display threshold does not establish that all retained
checkpoints implement the same computation.
Standing-suppression intervals use paired template-cluster
bootstrap resampling. The saved mean-trajectory data do not
provide an interval for the derived excess.
The two vertical axes have separate scales, held constant across
models, so the comparison concerns developmental profiles rather
than equality of magnitudes. The analysis measures association
across checkpoints and does not intervene on training.

\section{Readout Sensitivity}
\label{sec:app-readout-sensitivity}

These clean-run comparisons use the 2{,}000 held-out prompts of
Appendix~\ref{sec:app-backup-evaluation} to test how the measured
intermediate excess depends on the readout. The normalization
comparison shows the four models with intermediate overshoot;
the answer-score comparison includes all six final Pythia models.

\subsection{Normalization Convention}

\begin{figure*}[t]
  \centering
  \includegraphics[width=0.90\textwidth]{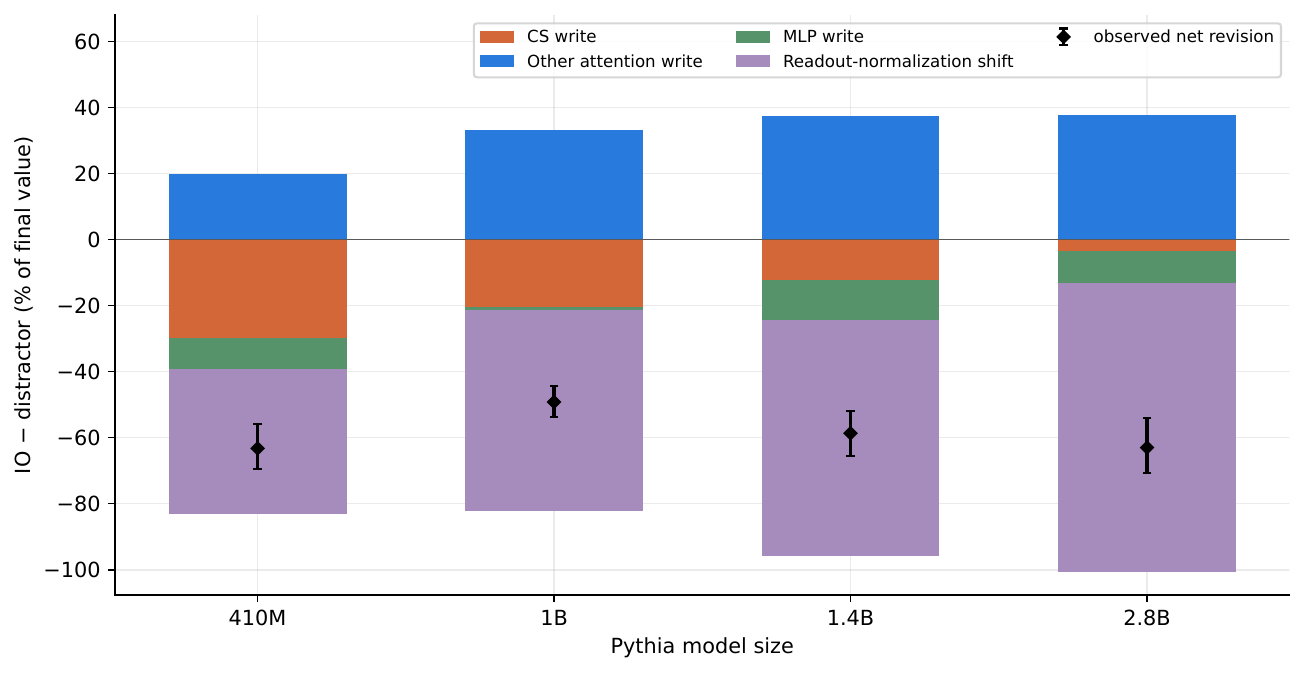}
  \caption{Revision from the intermediate peak to the final value
  of the mean layer-renormalized answer--distractor trajectory in
  four Pythia models. Bars show component writes under fixed
  clean-final normalization and the change in the difference between
  readout conventions; the latter is distinct from the repair term
  $N$. Diamonds show observed revision with 95\% paired
  template-cluster bootstrap intervals, reselecting the peak in each
  replicate. Values are percentages of each model's final
  answer--distractor logit difference. Pythia-6.9B and 12B are omitted
  because their mean trajectories peak at the final layer.}
  \label{fig:app-lens-normalization}
\end{figure*}

Figure~\ref{fig:app-lens-normalization} separates the revision
after the intermediate peak into component writes under fixed
clean-final normalization and a change between readout conventions.
It shows why the decline in the layer-renormalized trajectory
cannot be attributed to copy suppression alone.

\subsection{Answer-Score Definition}

\begin{figure*}[t]
  \centering
  \includegraphics[width=0.90\textwidth]{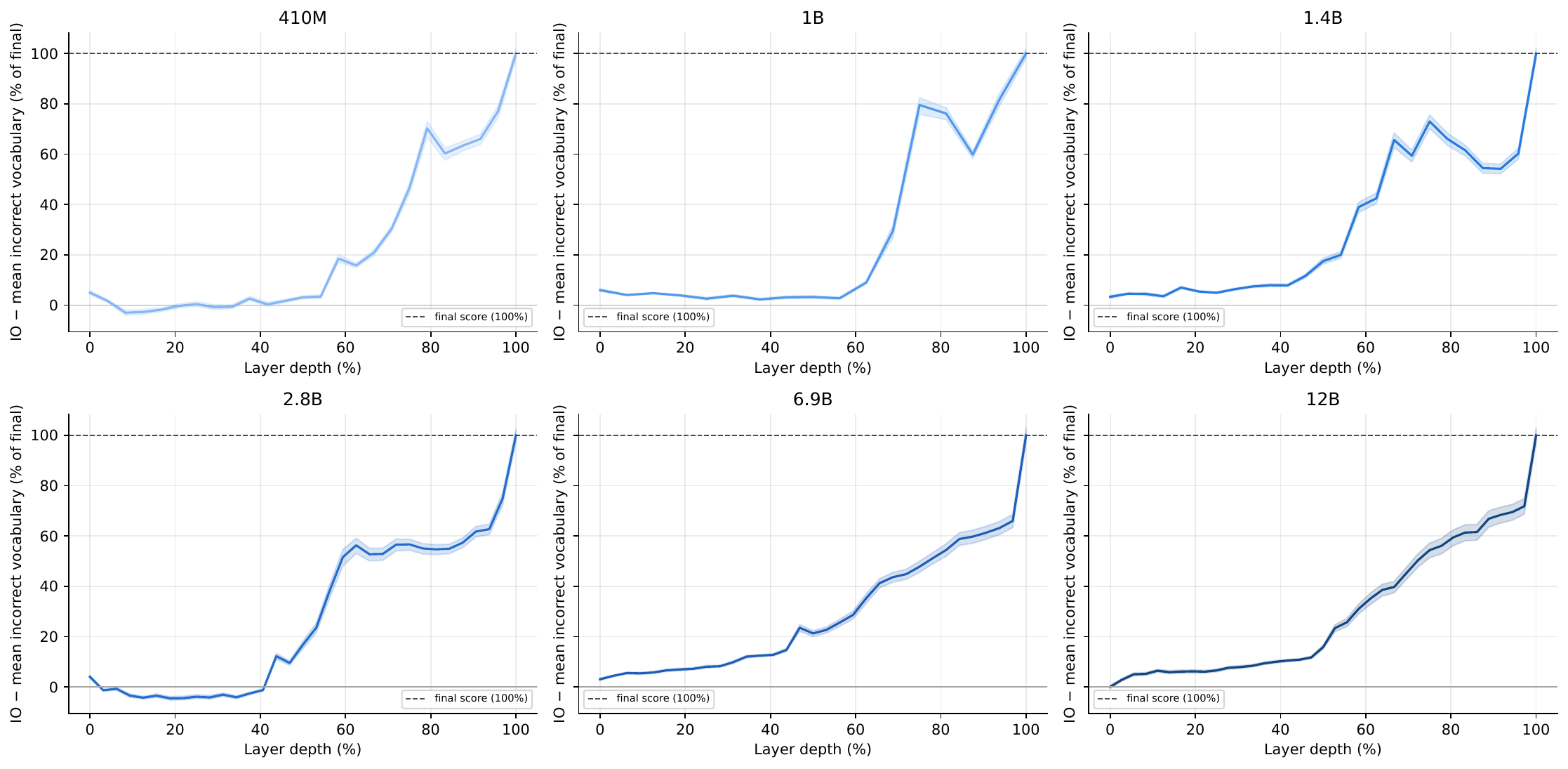}
  \caption{Mean clean-run answer scores with normalization
  recomputed at each layer, as a percentage of each model's final
  score. The score is the correct-answer
  logit minus the mean logit of all other vocabulary entries,
  including special and padded entries. Dashed lines mark the
  final value (100\%); shading gives 95\% paired template-cluster
  bootstrap intervals. These mean trajectories peak at the
  final layer; individual prompts may still exhibit
  intermediate declines.}
  \label{fig:app-lens-vocabulary}
\end{figure*}

Figure~\ref{fig:app-lens-vocabulary} replaces the distractor logit
with the mean logit of the other vocabulary entries. These mean
trajectories peak at the final layer, showing that intermediate
overshoot depends on the score being read out.

\end{document}